\PassOptionsToPackage{table}{xcolor}
\documentclass[]{aidata}

\usepackage{amsmath,amsfonts,bm}

\def\eqref#1{equation~\ref{#1}}

\def\1{\bm{1}}

\DeclareMathAlphabet{\mathsfit}{\encodingdefault}{\sfdefault}{m}{sl}
\SetMathAlphabet{\mathsfit}{bold}{\encodingdefault}{\sfdefault}{bx}{n}

\usepackage{lmodern}
\renewcommand{\sfdefault}{bytesans}
\usepackage{amssymb,booktabs,multirow,tabularx,url}
\usepackage{etoolbox,adjustbox,float}
\definecolor{OursRow}{RGB}{241,244,246}
\definecolor{PipeBlue}{RGB}{30,105,170}
\definecolor{PipeGreen}{RGB}{0,119,87}
\newcommand{\metricup}{\ensuremath{\uparrow}}
\newcommand{\metricdown}{\ensuremath{\downarrow}}
\newcommand{\ours}{RFW-R}
\newcommand{\NA}{--}
\AtBeginEnvironment{table}{\fontsize{8.5}{10}\selectfont\setlength{\tabcolsep}{3.5pt}}
\hypersetup{pdftitle={RFW-W: Scaling Physical Asset Universes for Generative Embodied Simulation},pdfauthor={},pdfcreator={},pdfproducer={}}

\title{RFW-W: Scaling Physical Asset Universes\\for Generative Embodied Simulation}

\author[1,*]{Jiawei Weng}
\author[1,2,*]{Rui Zhang}
\author[2,3,*,\dagger]{Saining Zhang}
\author[2,4,*]{Changjian Pan}
\author[1,*]{Yuanbo Lyu}
\author[1,2,*]{Jiarui Guo}
\author[3]{Yongzhi Li}
\author[2]{Guolin Liu}
\author[2]{Jialu Liu}
\author[2]{Jiankun Wang}
\author[1]{Yi Yang}
\author[3]{Peishuo Li}
\author[2]{Zhiheng Li}
\author[2]{Jiaying Wan}
\author[2]{Junhao Chen}
\author[5]{Weiqing Xiao}
\author[1,2]{Jingrun Du}
\author[1,2]{Wenyi Li}
\author[1]{Weiyi Lin}
\author[1,\ddagger]{Zenan Meng}
\author[1,\ddagger]{Hu Wei}
\author[1]{Lin Qu}
\author[2]{Zongzheng Zhang}
\author[2]{Hao Zhao}
\author[6,\ddagger]{Chaojian Li}

\affiliation[1]{Alibaba Group}
\affiliation[2]{Tsinghua University}
\affiliation[3]{Nanyang Technological University}
\affiliation[4]{Peking University}
\affiliation[5]{Nanjing University}
\affiliation[6]{Hong Kong University of Science and Technology}

\contribution[*]{Core contributors}
\contribution[\dagger]{Project leader}
\contribution[\ddagger]{Corresponding authors.}

\checkdata[Email]{\{mengzenan.mzn, kongwang\}@alibaba-inc.com, chaojian@ust.hk}
\checkdata[Project Page]{\url{https://roboflywheel-com.github.io/RoboFlywheel-World/}}

\abstract{Scaling embodied simulation requires diverse objects with executable physical representations. We present RFW-W, a physical asset universe connecting rigid, articulated, and deformable asset sources to task-conditioned scene composition. Its reported inventories comprise 1,010,386 asset records: 524,611 rigid assets in 1,212 categories (RFW-R), 402,410 articulated assets in 531 categories (RFW-A), and 83,365 jointly verified deformables in 485 categories (RFW-S). These taxonomies contribute 2,228 dataset--category entries. RFW-R couples category-guided generation with collision geometry, composed physical properties, and interaction annotations; RFW-A uses reusable agent-authored procedural generators with explicit part and joint structure; RFW-S uses structure-aware simulation representations, calibration, and independent testing. A common interface preserves physical requirements specific to each asset type for retrieval and scene assembly. Quantitative evaluations demonstrate that RFW-R provides broader asset coverage and stronger scene-fulfillment capabilities than existing datasets. Across five RoboTwin tasks, using 100 demonstrations per task for each asset source, policies trained on RFW-R achieve macro success rates of 71.2\% in clean settings and 41.8\% under randomization, exceeding the strongest baseline by 15.8 and 10.6 percentage points, respectively. RFW-W provides a scalable and unified asset foundation for generating physically grounded, task-conditioned embodied worlds.}

\begin{document}
\raggedbottom
\maketitle

\begin{figure}[htbp]
\centering
\includegraphics[width=0.99\linewidth]{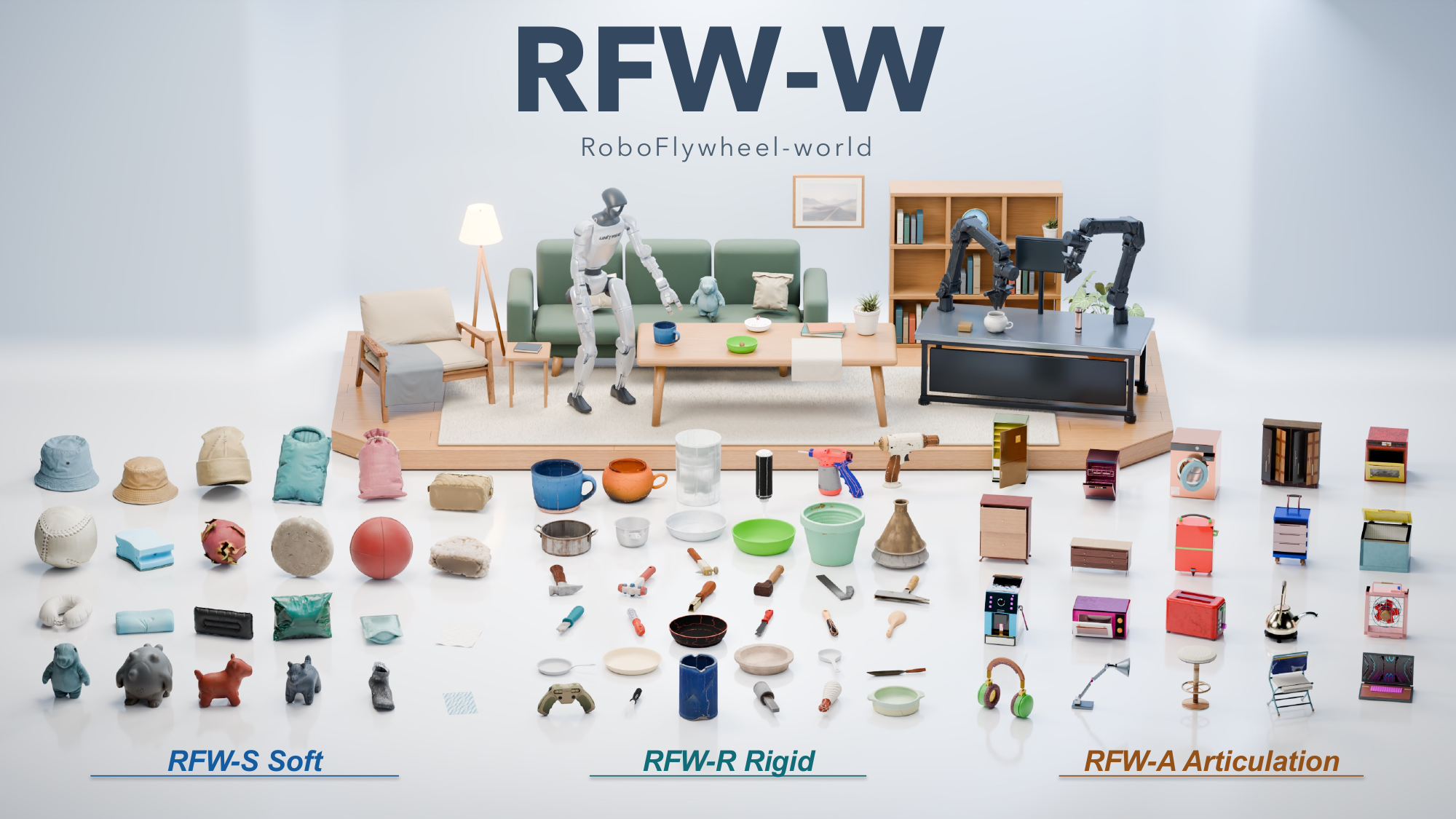}
\caption{\textbf{RFW-W physical asset universe.} RFW-W brings together three complementary asset sources: deformable objects (RFW-S, left), rigid objects (RFW-R, center), and articulated objects (RFW-A, right), connecting them through a common interface for task-conditioned embodied scene composition. Together, these sources comprise 1,010,386 reported asset records across 2,228 dataset–category entries. Each asset type retains its own construction and verification pipeline to preserve its specific physical properties and simulation requirements.}
\label{fig:teaser}
\end{figure}

%
%

\section{Introduction}
\label{sec:intro}
General-purpose embodied intelligence requires robots to operate across diverse objects, layouts, and interactions. Physics simulation provides a scalable alternative to real-world data collection, enabling reusable and varied environments for embodied learning~\citep{kolve2017ai2thor,procthor2022,robocasa2024,robogen2023,sage2026,embodiedgenv2_2026}. Yet its effective scope depends on the underlying asset library: scene generation and manipulation require not only diverse objects, but also representations that physics simulators can reliably execute.

Large-scale 3D repositories and generative models provide millions of assets and open-category synthesis of geometry and appearance~\citep{shapenet2015,objaverse2023,objaversexl2023,objaversepp2025,clay2024,trellis2024,trellis2_2025,step1x3d2025,hunyuan3d21_2025}. However, visual meshes alone do not specify metric scale, collision geometry, mass, friction, inertia, or interaction semantics. Collections such as YCB and Google Scanned Objects provide physical object models for manipulation and simulation~\citep{calli2015ycb,downs2022gso}, but extending these representations across broad categories and hundreds of thousands of assets remains challenging.

Recent physical-asset datasets and articulation-modeling methods enrich 3D content with physical, functional, and articulation information~\citep{physx3d2025,physxanything2025,physxomni2026,physforge2026,uniphysgen2026,iliash2026artiverse,yang2026sketch2arti}. ManiTwin adds manipulation-oriented digital twins, while Seed3D and EmbodiedGen connect assets to broader simulation-generation systems~\citep{manitwin2026,seed3d2_2026,embodiedgenv2_2026}. As shown in Table~\ref{tab:dataset-comparison}, existing embodied asset libraries have limited asset scale and category coverage, incomplete part-level physical annotations, and limited support for dexterous manipulation. These limitations motivate the development of a large-scale physical asset universe that unifies diverse object types with executable physical representations for embodied simulation.

\begin{table}[t]\fontsize{8.5}{10}\selectfont
\centering
\caption{\textbf{Comparison of embodied asset libraries.} R/A/D denotes rigid, articulated, and deformable objects, respectively; $\checkmark$ indicates support, $\times$ indicates no support, and $\triangle$ indicates partial support. \#Assets and \#Classes denote the numbers of assets and categories, while Med. inst. represents the median number of instances per category. Scene fill measures the fulfillment rate of 265 rigid-object requests across 16 fixed scene layouts. Seg. denotes part segmentation; Part phys. denotes numerical part-level physical properties with composed and verified mass, center of mass, and inertia; Hollow denotes hollow-aware mass modeling. Dexterous manip. refers to dexterous manipulation, where Hands indicates the number of supported dexterous hand models and Grasp ann. indicates grasp annotations. A dash (–) indicates unavailable results. RFW-W aggregates asset records and dataset-specific category entries from RFW-R, RFW-A, and RFW-S, while its scene-fulfillment and dexterous manipulation results are inherited from RFW-R.}
\label{tab:dataset-comparison}
\begin{adjustbox}{max width=\linewidth}
\begin{tabular}{@{}l c rrrr ccc cc@{}}
\toprule
\multirow{2}{*}{Dataset} & \multirow{2}{*}{R/A/D} & \multicolumn{4}{c}{Coverage and scene use} & \multicolumn{3}{c}{Part-level physics} & \multicolumn{2}{c}{\shortstack{Dexterous\\manip.}} \\
\cmidrule(lr){3-6}\cmidrule(lr){7-9}\cmidrule(lr){10-11}
 & & \#Assets & \#Classes & \shortstack{Med.\\inst.} & \shortstack{Scene\\fill} & Seg. & \shortstack{Part\\phys.} & Hollow & Hands & \shortstack{Grasp\\ann.} \\
\midrule
EmbodiedGen v2~\citep{embodiedgenv2_2026} & $\checkmark$/$\times$/$\triangle$ & 4,168 & 200 & 11 & 97.4\% & $\checkmark$ & $\times$ & $\times$ & 0 & $\times$ \\
PhysX-3D~\citep{physx3d2025} & $\checkmark$/$\triangle$/$\times$ & 32,047 & 450 & 1 & \NA & $\checkmark$ & density & $\times$ & 0 & $\times$ \\
ManiTwin~\citep{manitwin2026} & $\checkmark$/$\times$/$\times$ & 4,574 & 485 & 6 & 90.6\% & $\times$ & $\times$ & $\times$ & 0 & $\checkmark$ \\
\midrule
\rowcolor{OursRow} RFW-R & $\checkmark$/$\times$/$\times$ & \textbf{524,611} & \textbf{1,212} & \textbf{356} & \textbf{100\%} & $\checkmark$ & $\checkmark$ & $\checkmark$ & 6 & $\checkmark$ \\
\rowcolor{OursRow} RFW-W & $\checkmark$/$\checkmark$/$\checkmark$ & \textbf{1,010,386} & \textbf{2,228} & \textbf{438} & \textbf{100\%} & $\checkmark$ & $\checkmark$ & $\checkmark$ & 6 & $\checkmark$ \\
\bottomrule
\end{tabular}\end{adjustbox}
\end{table}


We introduce \textbf{RoboFlywheel-World (RFW-W)}, a large-scale physical asset universe for embodied simulation, covering rigid, articulated, and deformable objects. RFW-W comprises three asset collections: \textbf{RFW-R}, containing 524,611 rigid assets across 1,212 categories; \textbf{RFW-A}, containing 402,410 articulated assets across 531 categories~\citep{rfwa2026}; and \textbf{RFW-S}, containing 83,365 jointly verified deformable assets across 485 categories~\citep{rfws2026}. Together, these collections comprise \textbf{1,010,386 asset records} across 2,228 dataset--category entries.

RFW-W organizes these assets through type-specific construction pipelines. RFW-R, the focus of this work, combines category-guided generation with part segmentation, collision geometry, part-level physical properties, and simulation-validated grasp annotations across six dexterous hands. RFW-A employs agent-authored procedural generators to create articulated assets with explicit part and joint structures~\citep{rfwa2026}, while RFW-S uses structure-aware simulation representations, calibration, and independent physical verification~\citep{rfws2026}. A common interface connects all three collections to task-conditioned asset retrieval and scene composition.


We conduct extensive evaluations to assess the coverage, physical fidelity, and downstream utility of RFW-R. Our evaluations cover asset and part-level coverage, physical annotation quality, cross-simulator consistency, dexterous grasp execution, scene construction, and policy learning. In scene construction, RFW-R fulfills all 265 object requests across 16 fixed scene foundations (Table~\ref{tab:dataset-comparison}). Furthermore, policies trained on RFW-R achieve the strongest macro success rates across five RoboTwin tasks under matched demonstration budgets (Table~\ref{tab:policy-transfer}). These results demonstrate the effectiveness of RFW-R in supporting physically grounded scene construction and downstream robotic manipulation.

Our contributions are summarized as follows:

\begin{itemize}
    \item We present \textbf{RFW-W}, a large-scale physical asset universe comprising 1,010,386 asset records across rigid, articulated, and deformable objects, organized through a unified interface for task-conditioned asset retrieval and scene composition.

    \item We develop \textbf{RFW-R}, a large-scale rigid asset collection featuring category-guided generation, part-level semantic and physical annotations, collision geometry, and simulation-validated grasp annotations across six dexterous hands.

    \item We conduct comprehensive evaluations of RFW-R, covering asset and part-level coverage, physical annotation quality, cross-simulator consistency, dexterous grasp execution, scene fulfillment, and downstream policy learning across five RoboTwin tasks.
\end{itemize}
%

\section{Related Work}
\label{sec:related}
\subsection{3D Asset Collections and Physical Asset Datasets}

Large-scale 3D repositories, including ShapeNet, 3D-FUTURE, ABO, OmniObject3D, Objaverse, Objaverse-XL, and Objaverse++, have substantially expanded the scale, category coverage, and visual diversity of available 3D assets~\citep{shapenet2015,fu2020future,abo2021,omniobject3d2023,objaverse2023,objaversexl2023,objaversepp2025}. Recent generative models, such as CLAY, TRELLIS, TRELLIS.2, Hunyuan3D 2.1, and etc., further enable open-category synthesis of high-quality geometry, textures, and PBR materials~\citep{clay2024,trellis2024,trellis2_2025,hunyuan3d21_2025,step1x3d2025,hi3dgen2025,idea23d2025,chen2026unigarment}. These resources primarily focus on visual appearance and geometric reconstruction, leaving the physical properties and interaction capabilities required for embodied simulation largely unaddressed.

Early manipulation-oriented resources, such as YCB, provide physical object models for robotic manipulation~\citep{calli2015ycb}, while Google Scanned Objects provides scanned household assets for simulation. More recent physical asset datasets extend this direction by incorporating metric scale, material properties, physical attributes, affordances, and simulation-ready representations~\citep{physx3d2025,physxanything2025,physforge2026,uniphysgen2026}. ManiTwin introduces large-scale manipulation-oriented digital assets with grasp annotations and verified interaction proposals~\citep{manitwin2026}, while Seed3D and EmbodiedGen further connect physical asset generation to broader embodied simulation and world-generation systems~\citep{seed3d1_2025,seed3d2_2026,embodiedgen2025,embodiedgenv2_2026}.

As summarized in Table~\ref{tab:dataset-comparison}, existing physical asset libraries vary substantially in asset scale and category coverage. Moreover, detailed part-level physical properties and dexterous manipulation support are not consistently available across these datasets. RFW-W addresses these limitations by organizing rigid, articulated, and deformable assets within a unified framework for embodied simulation. It comprises RFW-R for large-scale rigid assets, RFW-A for agent-authored procedural generation of articulated objects~\citep{rfwa2026}, and RFW-S for everyday deformable assets with configuration-specific physical and runtime verification~\citep{rfws2026}.

\subsection{Physicalization and Interaction-Ready Asset Representations}

Semantic and part-level representations provide the structural foundation for physical asset construction. Shape-aware zero-shot semantic segmentation combines semantic recognition with spatial delineation using spectral cues from self-supervised pixel features~\citep{liu2023shapeaware}. PartNet establishes fine-grained hierarchical part annotations for 3D objects~\citep{partnet2019}, while SAPIEN provides a part-based interactive simulation environment for articulated objects~\citep{Xiang_2020_SAPIEN}. GAPartNet further studies generalizable and actionable object parts across categories, connecting part perception to robotic manipulation~\citep{gapartnet2023}. These works establish the importance of identifying meaningful object parts beyond holistic geometric representations.

Beyond semantic structure, physicalization requires assigning physical properties and constructing collision representations that can be executed by physics simulators. CoACD generates convex collision decompositions while accounting for geometric concavity~\citep{coacd2022}. PUGS combines shape-aware Gaussian reconstruction, vision-language property inference, feature-based propagation, and volume integration for zero-shot physical understanding and object-mass estimation~\citep{pugs2025}. Building on these directions, RFW-R assigns material and friction properties to semantic parts, estimates their physical attributes, and composes part-level mass, center of mass, and inertia into explicit rigid-body representations. It further constructs convex collision geometry and preserves the correspondence between visual geometry, collision geometry, and physical properties.

Interaction-ready assets additionally require annotations and executable grasp representations. ACRONYM and DexGraspNet provide large-scale simulated grasp datasets for general objects~\citep{acronym2020,dexgraspnet2022}, while GraspXL studies grasp motion generation for diverse objects~\citep{graspxl2024}. ManiTwin and EmbodiedGen further connect physical asset annotations with manipulation-oriented interactions and embodied simulation~\citep{manitwin2026,embodiedgenv2_2026}. However, grasp annotations or generated grasp proposals alone do not guarantee successful execution in simulation. RFW-R connects semantic part representations to dexterous grasp generation and validates candidate grasps through contact, penetration, closing, lifting, holding, and perturbation checks across six dexterous hand models. Its physical representations are also evaluated across multiple physics simulators to assess the consistency of their simulated behavior.

\subsection{From Physical Assets to Generative Embodied Worlds}

Procedural and language-driven scene generation has enabled increasingly diverse and scalable embodied environments through structured generation, asset retrieval, learned layouts, and physical constraints~\citep{procthor2022,infinigen2023,infinigenindoors2024,layoutgpt2023,holodeck2023,anyhome2023,physcene2024,instructscene2024,xu2025diffusionknowstransparency}. Related work has further explored interactive environments, scene understanding, controllable scene editing, and the reconstruction of simulation-ready scenes from real-world observations~\citep{kolve2017ai2thor,fu2020front,pqtransformer2022,sceneeditor2026,wang2026tableverse,shuai2026gennav}. More recent agentic approaches incorporate multimodal reasoning, iterative planning, tool use, and visual or simulation feedback to construct task-conditioned, editable, and physically executable worlds~\citep{scenethesis2025,sceneweaver2025,scenerevis2026,sage2026,embodiedgenv2_2026,lucida2026,neoworldpro2026}. In parallel, simulation platforms and data engines scale environments, tasks, and demonstrations for robot learning~\citep{habitat2_2021,habitat3_2023,igibson2_2021,behavior1k2024,maniskill2_2023,maniskill3_2024,robocasa2024,robotwin2_2025,mimicgen2023,gensim2023,robogen2023}. While these efforts primarily focus on generating, reconstructing, and scaling downstream scenes and interaction data, RFW-W focuses on the upstream physical asset foundation, organizing rigid, articulated, and deformable assets through a unified retrieval interface and connecting them to task-conditioned scene composition and downstream robotic learning.

\subsection{Robot Policies and Training Data}

Diffusion Policy models visuomotor actions with a conditional diffusion process~\citep{chi2023diffusionpolicy}, while RT-2 connects vision-language pretraining to robotic action prediction~\citep{brohan2023rt2}. Open X-Embodiment aggregates demonstrations across robot platforms~\citep{openxembodiment2023}, and Octo studies a generalist policy trained on this data and adapted to new settings~\citep{octomodelteam2024octo}. These works motivate scalable policy learning from diverse experience. Our policy experiment examines a different input to that experience: the physical asset library used to collect task demonstrations. It holds the policy family and demonstration count fixed across libraries, while differences in category match, trajectories, and asset preparation remain part of the training-source comparison.

\section{Method}\label{sec:method}
RFW-W draws on rigid, articulated, and deformable asset sources for task-conditioned retrieval and scene construction (Figures~\ref{fig:teaser} and~\ref{fig:scene}). These sources retain distinct physical representations and verification records; a common interface exposes the fields required for a task while retaining the simulation models appropriate to each asset type.

\subsection{Physical asset sources and construction procedures}\label{sec:asset-universe}
\paragraph{Rigid assets: category coverage and executable interactions.}
RFW-R supplies 524,611 retained objects across 1,212 leaf categories. Category-conditioned generation and multi-view inspection establish coverage; semantic parts, convex collisions, composed physical properties, and interaction annotations make the assets usable for rigid-body simulation. Sections~\ref{sec:asset-generation} and~\ref{sec:physics-annotation} describe this construction, and Section~\ref{sec:experiments} evaluates its asset, grasp, scene, and policy outcomes.

\paragraph{Articulated assets: reusable programs and motion-aware checks.}
RFW-A supplies 402,410 articulated assets across 531 categories~\citep{rfwa2026}. Agents organize reusable components and their dependencies into category blueprints, implement executable generators, and repair failures across sampled configurations and joint states. Validated generators then sample new assets without per-asset model inference. Each sample retains its part hierarchy, joint parameters, visual and collision geometry, physical-property annotations, generator version, and seed. The companion study reports a 100\% asset-level kinematic pass rate over its release and 99.5\% passage of a one-second, root-fixed passive rollout in all four tested simulators on a 200-asset subset. These checks concern articulated structure and the stated passive regime; they are distinct from the RFW-R drop-response protocol. Articulation-estimation and real-object manipulation studies provide additional evidence for articulated assets.

\paragraph{Deformable assets: structural adaptation and independent testing.}
RFW-S contributes 83,365 jointly verified assets across 485 categories: soft solids, cloth sheets, shells, garments, and bags~\citep{rfws2026}. Structure-aware adapters select compatible position-based dynamics (PBD) or material point method (MPM) representations and map inferred parameters to controls actually executed in Genesis 1.3.3. Calibration supports bounded corrections; a different physical probe tests the frozen candidate, and a separate check measures runtime behavior. The configuration-matched intersection retains 83,365 assets and all 485 categories. The resulting assets satisfy both physical-response and runtime criteria under the evaluated simulation configurations. Additional garment and grasping studies evaluate deformation behavior, manipulation performance, and object integrity.

\paragraph{Unified Asset Organization and Retrieval.}
RFW-W integrates rigid, articulated, and deformable assets through a unified interface for task-driven retrieval and scene construction. The interface organizes semantic, geometric, physical, and interaction information while incorporating type-specific representations, including collision geometry and grasp annotations for rigid objects, part hierarchies and joint parameters for articulated objects, and deformation models and simulation parameters for deformable objects. This unified organization enables efficient asset retrieval and scene composition based on task requirements and physical compatibility.

\subsection{Category-guided Asset Generation}
\label{sec:asset-generation}

\begin{figure}[t]
\centering
\includegraphics[width=\textwidth]{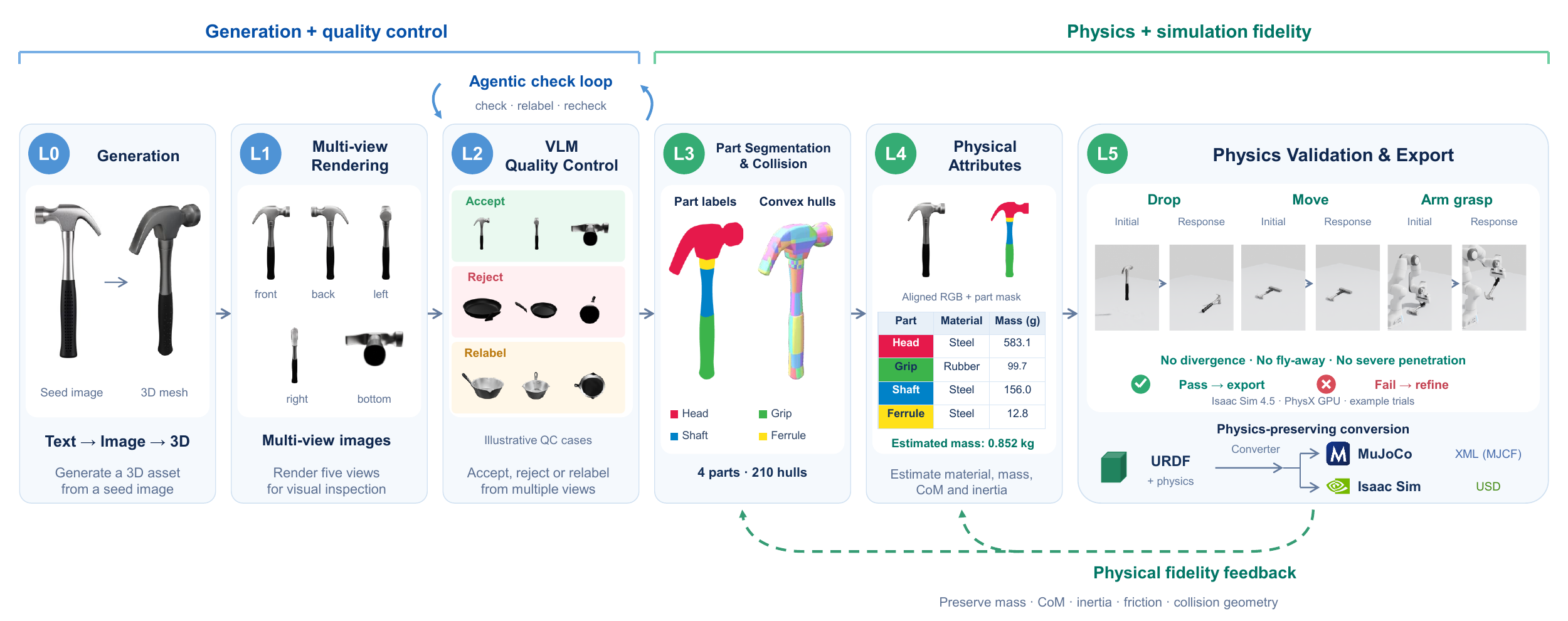}
\caption{\textbf{RFW-R rigid-asset construction pipeline.} Stages L0--L2 perform category-guided generation, multi-view rendering, and VLM-based quality control. Stages L3--L4 construct semantic part and collision representations, estimate physical attributes, and export the physicalized assets; L5 summarizes downstream simulation evaluations. The blue feedback loop supports category reassignment and reinspection, while the green feedback loop guides physical refinement. Assets exported after L4 support URDF, MJCF, and USD representations.}
\label{fig:pipeline}
\end{figure}

Figure~\ref{fig:pipeline} presents the RFW-R construction pipeline, spanning asset generation and quality control (L0--L2), physicalization and export (L3--L4), and downstream simulation evaluation (L5). In L0, we employ a human-curated taxonomy to guide large-scale asset generation. Qwen3.7-plus generates diverse category-conditioned object descriptions, which are converted into images using FLUX.2-klein~\citep{blackforestlabs2026fluxklein} and a supplementary Wan2.7 generation path. RMBG-2.0~\citep{bria2024rmbg} extracts object foregrounds, and TRELLIS.2~\citep{trellis2_2025} reconstructs textured 3D meshes.

In L1, each generated asset is rendered from five viewpoints (front, back, left, right, and bottom). In L2, Qwen3.7-plus~\citep{qwen3,qwen37plus} evaluates the multi-view renderings for geometric completeness, object recognizability, rigid-body suitability, and category consistency. An agentic quality-control loop accepts qualified assets, rejects unsuitable candidates, or reassigns incorrectly classified objects followed by reinspection. This generation and filtering process yields 524,611 retained rigid assets across 1,212 categories. Additional implementation details and prompt specifications are provided in Appendix~\ref{app:repair}.

\subsection{Physicalization and Interaction Annotation}
\label{sec:physics-annotation}

Following visual quality control, L3 constructs semantic part and collision representations. P3-SAM~\citep{p3sam2025} identifies functional parts, while CoACD~\citep{coacd2022} decomposes object geometry into convex collision components for physics simulation. In L4, Qwen3.7-plus analyzes aligned RGB images and part masks to estimate part materials, friction coefficients, metric dimensions, and whole-object mass priors. Each part is assigned a mass $m_p=\rho_p V_p$ based on its estimated material density and volume. Part-level properties are then composed into the object's total mass, center of mass, and inertia tensor. When the ratio between the composed mass and the whole-object mass estimate falls outside $[0.5,2]$, the part masses are uniformly rescaled while preserving their relative proportions.

After completing L4 physicalization, assets are exported to URDF and converted into MJCF and USD representations, preserving their geometric and physical properties across simulation formats. L5 denotes downstream simulation evaluations of the exported assets, with task-specific protocols and results reported in the experiments and appendices. These evaluations characterize asset behavior and provide feedback for physical refinement. Detailed physical annotation and composition procedures are described in Appendices~\ref{app:repair} and~\ref{app:physics}.

To further support robotic manipulation, we generate part-conditioned dexterous grasp annotations for six robotic hand models. Building on TRO-Grasp~\citep{fei2025trograspefficientgraph}, we condition grasp generation on target semantic parts through partial point clouds while retaining its hand-conditioned pose representation. Candidate grasps are filtered through semantic-part contact, mesh penetration, multi-finger engagement, and simulated closing, lifting, holding, and perturbation checks. The same annotation and filtering protocol is applied across the supported hands, producing executable grasp annotations associated with each asset's semantic and physical representations. Additional details are provided in Appendix~\ref{app:grasp}.

\subsection{Task-driven Scene Construction}
\label{sec:scene}

Building on the constructed physical asset collections, RFW-W provides a unified interface for task-conditioned asset retrieval and scene generation. The interface organizes semantic, geometric, physical, and interaction information while preserving the representations required by rigid, articulated, and deformable objects. Given a task description, a language model translates the requested objects and their spatial relationships into a structured scene specification. Assets are then retrieved from the corresponding collections based on semantic relevance and physical compatibility.

\begin{figure}[t]
\centering
\includegraphics[width=1.0\linewidth]{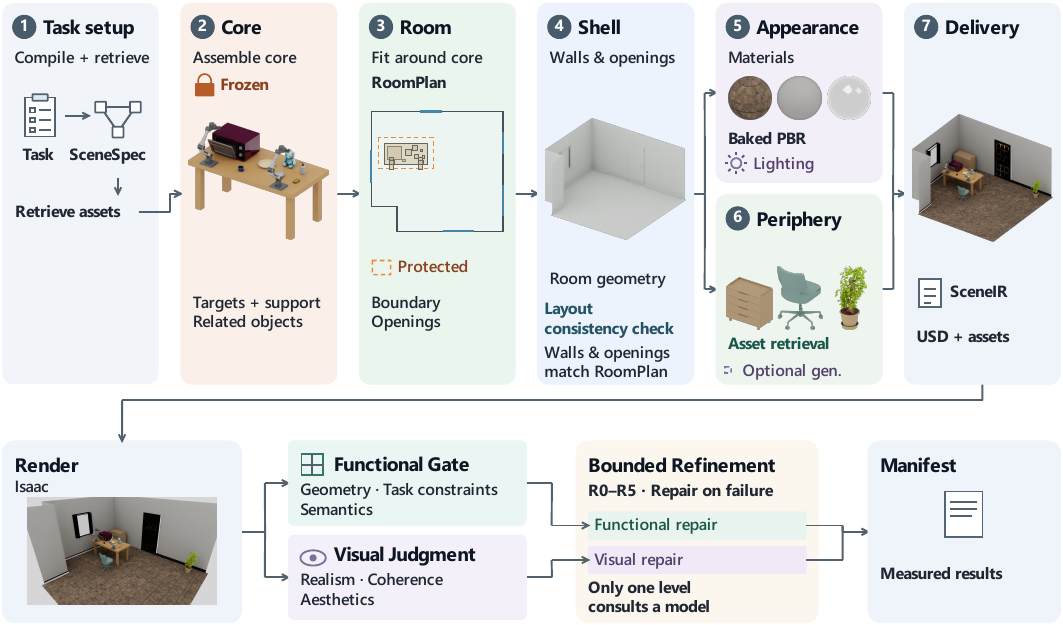}
\caption{\textbf{Task-driven scene construction.} Task instructions are compiled into structured specifications, followed by asset retrieval and task-core assembly. The frozen core guides room generation, appearance synthesis, and scene composition, while functional and visual checks support bounded refinement before export.}
\label{fig:scene}
\end{figure}

As illustrated in Figure~\ref{fig:scene}, scene construction follows a task-core-first strategy. A geometric solver first assembles the task-relevant objects, including targets, supporting objects, and their required spatial relationships, in a room-independent coordinate frame. Once the task core is established, its object membership and relative poses are frozen. The surrounding room is then constructed using Infinigen~\citep{infinigen2023,infinigenindoors2024}, which generates walls, floors, doors, and windows while respecting the frozen task core. Peripheral furniture, materials, and lighting are subsequently added under protected-region constraints, with procedural materials baked into PBR textures.

\begin{figure}[t]
\centering
\includegraphics[width=\linewidth]{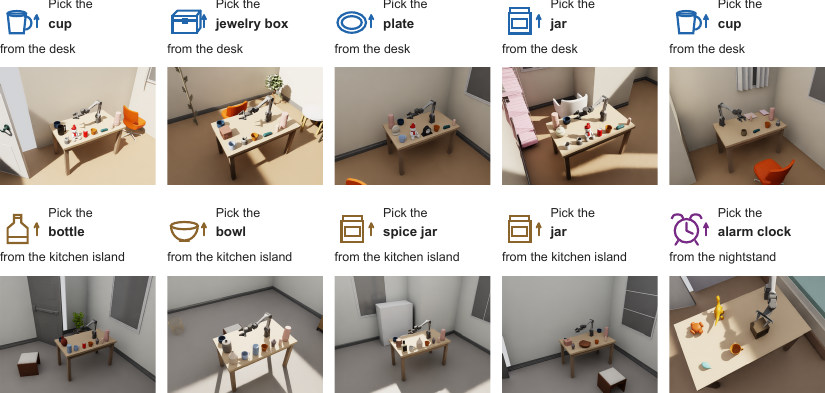}
\caption{\textbf{Task-aligned scene examples.} Ten task instructions paired with their corresponding scene renderings. Each example illustrates the requested action, target object, and support surface in the generated scene.}
\label{fig:scene-multisim}
\end{figure}

The generated scene undergoes functional and visual validation before export. Deterministic checks evaluate geometric consistency, task constraints, and semantic correctness, while a Qwen3.7-plus vision-language judge assesses visual realism and scene coherence. Failed checks trigger bounded refinement through separate functional and visual repair channels (R0--R5), without modifying the frozen task core. The final scene is exported as a structured description and self-contained USD package for Isaac Sim~\citep{NVIDIA_Isaac_Sim}. Figure~\ref{fig:scene-multisim} presents task-aligned examples across different indoor environments. The scene-fulfillment and downstream policy evaluations in this work are conducted using RFW-R assets.

\section{Experiments}
\label{sec:experiments}
We evaluate RFW-R from asset construction to downstream use, examining semantic and physical annotation quality, cross-simulator behavior, dexterous grasp execution, robot policy learning, and task-driven scene construction. Cross-library comparisons use RFW-R as the rigid-asset source, while RFW-A and RFW-S are evaluated in their companion studies. The evaluation cohorts, metrics, and protocols are detailed in Appendix~\ref{app:release}.

\subsection{Asset annotation and physical consistency}
\label{sec:exp-asset}\label{sec:exp-geometry}
We assess whether RFW-R provides reliable semantic annotations and physically consistent asset descriptions. Under the shared vision-language evaluation protocol, RFW-R achieves the highest category agreement (0.981) and functional agreement (0.860) among the compared libraries; its language agreement (0.864) is close to PhysX-3D (0.870), as shown in Table~\ref{tab:annotation-plan}. Metric definitions and object-level diagnostics are provided in Appendix~\ref{app:audit}.

\begin{table}[htbp]\fontsize{8.5}{10}\selectfont
\caption{Part-level semantic annotation agreement evaluated with a shared vision-language judging protocol. Category, Language, and Functional denote the three annotation dimensions; higher scores indicate stronger agreement. Additional diagnostics are given in Appendix~\ref{app:audit}.}
\label{tab:geometry}\label{tab:annotation-plan}
\centering
\begin{adjustbox}{max width=\linewidth}\begin{tabular}{lrrrr}
\toprule
Measure & \ours{} & EmbodiedGen v2 & PhysX-3D & ManiTwin \\
\midrule
Category \metricup & \textbf{0.981} & 0.93 & 0.96 & 0.65 \\
Language \metricup & 0.864 & 0.67 & \textbf{0.87} & 0.50 \\
Functional \metricup & \textbf{0.860} & 0.80 & 0.79 & 0.64 \\
\bottomrule
\end{tabular}\end{adjustbox}
\end{table}

\paragraph{Physical representation.}\label{sec:exp-physics}
We examine physical-parameter consistency on 2,424 assets. The exported URDF and MJCF masses match the assigned object masses with a maximum relative error of $1.5\!\times\!10^{-6}$, and 98.5\% of assets satisfy a relative tolerance of $10^{-3}$ between the sum of part masses and the stored object mass. Additional center-of-mass and inertia checks are reported in Appendix~\ref{app:physics}.

\paragraph{Effect of solid-fill modeling.}\label{sec:exp-physical-ablation}
We further compare the native physical representations against solid-fill modeling on 31 retained assets. Replacing the native representation with solid fill decreases the mean drop-response criterion-pass fraction by 9.25, 9.25, and 7.53 percentage points in MuJoCo, Genesis, and Isaac Sim, respectively. The consistent decrease across engines highlights the importance of the mass distribution encoded by the native representations.

\subsection{Cross-simulator drop-response consistency}
\label{sec:exp-crosssim}
To examine how the same physical assets behave across simulation engines, we conduct standardized drop-response tests on 2,257 common assets in PyBullet~\citep{coumans2016pybullet}, MuJoCo~\citep{todorov2012mujoco}, Genesis~\citep{Genesis}, and Isaac Sim~\citep{NVIDIA_Isaac_Sim}. All engines use the same estimated physical properties and multi-hull collision geometry. Each asset is tested at three drop heights and five initial orientations. We calculate the fraction of trials satisfying the shared displacement, rotation, and penetration criteria, binarize it at 0.5 for agreement analysis, and use Pearson $r$ to compare the continuous fractions. The evaluation criteria are specified in Appendix~\ref{app:crosssim}.

Across the four engines, 80.77\% of assets receive the same thresholded label (Fleiss' $\kappa=0.481$), counting both shared passes and shared failures. MuJoCo and Genesis show the highest pairwise agreement (94.20\%, Cohen's $\kappa=0.741$) and the strongest correlation in per-asset criterion-pass fractions ($r=0.919$; Table~\ref{tab:crosssim}). These measurements quantify the consistency of RFW-R's physical behavior across simulation backends.

\begin{table}[htbp]\fontsize{8.5}{10}\selectfont
\caption{Cross-simulator drop-response agreement on 2,257 common assets. Agreement and Cohen's $\kappa$ are computed from criterion-pass fractions binarized at 0.5, whereas Pearson $r$ compares continuous fractions. All four engines agree on 80.77\% of assets (Fleiss' $\kappa=0.481$).}
\label{tab:crosssim}
\centering

\begin{adjustbox}{max width=\linewidth}\begin{tabular}{lrrr}
\toprule
Engine pair & Agreement (\%) \metricup & Cohen's $\kappa$ \metricup & Pearson $r$ \metricup \\
\midrule
PyBullet--MuJoCo & 87.55 & 0.430 & 0.688 \\
PyBullet--Genesis & 86.71 & 0.398 & 0.670 \\
PyBullet--Isaac Sim & 87.90 & 0.331 & 0.650 \\
MuJoCo--Genesis & 94.20 & 0.741 & 0.919 \\
MuJoCo--Isaac Sim & 90.87 & 0.507 & 0.838 \\
Genesis--Isaac Sim & 89.68 & 0.450 & 0.830 \\
\bottomrule
\end{tabular}\end{adjustbox}
\end{table}
\subsection{Dexterous grasp execution}
\label{sec:exp-manipulation}
We evaluate whether the part-conditioned grasp annotations can be executed by six dexterous hand models in Genesis. For each object--hand pair, up to 64 grasp candidates undergo closing, lifting, holding, and six-direction perturbation tests. A lift is successful when an object rises more than 0.075\,m following a 0.15\,m commanded motion. Robust success additionally requires successful holding and perturbation under three mass, friction, and pose settings. We report object-level any-success coverage: an object counts as successful if at least one evaluated candidate passes the corresponding test.

\begin{table}[htbp]\fontsize{8.5}{10}\selectfont
\caption{Simulated dexterous grasp outcomes across six hand models. $N_L$ and $N_R$ are the numbers of evaluated objects for lift and robust success, respectively; the two metrics use different object cohorts. Macro denotes the unweighted mean across hands. Confidence intervals are reported in Appendix~\ref{app:grasp}.}
\label{tab:grasp-quality}
\centering
\begin{adjustbox}{max width=\linewidth}\begin{tabular}{lrrrr}
\toprule
Hand & $N_L$ & Lift (\%) \metricup & $N_R$ & Robust (\%) \metricup \\
\midrule
Ability & 2,372 & 92.5 & 2,367 & 90.4 \\
Inspire & 2,334 & 92.3 & 2,353 & 90.4 \\
Allegro & 2,372 & 91.3 & 2,372 & 81.6 \\
Barrett & 2,353 & 84.4 & 2,372 & 81.5 \\
XHand & 2,315 & 74.1 & 2,315 & 71.2 \\
Shadow & 2,353 & 42.3 & 2,372 & 27.3 \\
\midrule
Macro & & 79.5 & & 73.7 \\
\bottomrule
\end{tabular}\end{adjustbox}
\par\medskip
\normalsize
\centering
\includegraphics[width=0.90\textwidth]{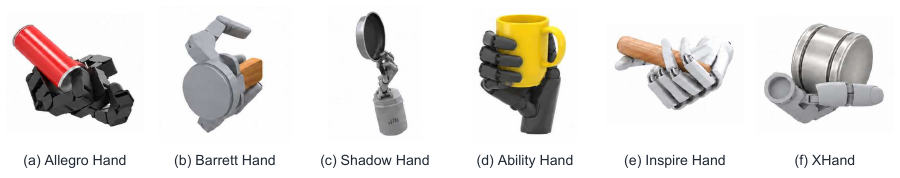}
\captionof{figure}{\textbf{Dexterous grasp execution across six hand models.} Representative successful grasp-and-lift examples using RFW-R assets, evaluated in Genesis under the stated lift criterion.}
\label{fig:dexterous-qual}
\end{table}

With 64 candidates per object, the macro object-level coverage reaches 79.5\% for lifting and 73.7\% for robust grasping (Table~\ref{tab:grasp-quality}). Ability and Inspire exceed 90\% on both metrics, whereas Shadow reaches 42.3\% lift coverage and 27.3\% robust coverage. Figure~\ref{fig:dexterous-qual} illustrates successful examples across the six hand morphologies. The variation between hands emphasizes the need for execution-based grasp assessment rather than relying on generated poses alone; candidate-budget and contact analyses appear in Appendix~\ref{app:grasp}.

\subsection{Policy learning and real-robot trajectory replay}
\label{sec:exp-policy-transfer}
We assess whether demonstrations collected with RFW-R assets improve downstream manipulation policies. We train separate $\pi_{0.5}$~\citep{physicalintelligence2025pi05} policies for each of four asset libraries across five RoboTwin~\citep{robotwin2_2025} tasks, using 100 demonstrations per library--task pair and the checkpoint at training step 19,999. Each policy is tested on 100 clean and 100 randomized expert-solvable scenes, resulting in 4,000 closed-loop simulation rollouts. Randomization varies visual context, table height, and language instructions; full training and evaluation settings appear in Appendix~\ref{app:policy}.

\begin{table}[htbp]\fontsize{8.5}{10}\selectfont
\caption{Closed-loop RoboTwin policy success in simulation (\%). Each asset library provides 100 training demonstrations per task; each task--condition cell is evaluated on 100 expert-solvable scenes. Clean and Randomized denote the two evaluation conditions, and Macro is the unweighted average across the five tasks.}
\label{tab:policy-transfer}
\centering

\begin{adjustbox}{max width=\linewidth}
\begin{tabular}{llrrrrrr}
\toprule
Mode & Training assets & A2B-L \metricup & Stand \metricup & Plate \metricup & Bowls-2 \metricup & Pillbottle-Pad \metricup & Macro \metricup \\
\midrule
\rowcolor{OursRow} Clean & \ours{} & \textbf{47} & \textbf{63} & \textbf{96} & \textbf{98} & \textbf{52} & \textbf{71.2} \\
 & EmbodiedGen v2 & 3 & 0 & 24 & 0 & 0 & 5.4 \\
 & PhysX-3D & 22 & \underline{51} & \underline{93} & 42 & 6 & 42.8 \\
 & ManiTwin & \underline{33} & 50 & 81 & \underline{80} & \underline{33} & \underline{55.4} \\
\midrule
\rowcolor{OursRow} Randomized & \ours{} & \textbf{21} & \textbf{33} & \textbf{76} & \textbf{73} & \underline{6} & \textbf{41.8} \\
 & EmbodiedGen v2 & 1 & 0 & 10 & 0 & 0 & 2.2 \\
 & PhysX-3D & \underline{8} & 16 & 48 & 24 & \textbf{9} & 21.0 \\
 & ManiTwin & \underline{8} & \underline{29} & \underline{56} & \underline{59} & 4 & \underline{31.2} \\
\bottomrule
\end{tabular}\end{adjustbox}
\end{table}
RFW-R-trained policies achieve the highest macro success rates in both clean (71.2\%) and randomized (41.8\%) evaluations, improving on the strongest baseline, ManiTwin, by 15.8 and 10.6 percentage points, respectively (Table~\ref{tab:policy-transfer}). RFW-R leads on all five clean tasks and four of the five randomized tasks; PhysX-3D performs better on randomized Pillbottle-Pad (9\% versus 6\%). The matched demonstration budgets show the downstream value of the RFW-R training source under the evaluated task settings.

\begin{figure}[t]
\centering
\begingroup
\setlength{\fboxsep}{5pt}
\colorbox{OursRow}{\parbox{0.92\linewidth}{\centering\small
Simulation: policy inference $\longrightarrow$ Recorded trajectory $\longrightarrow$ Real robot: replay}}
\endgroup
\par\smallskip
\includegraphics[width=1.0\linewidth]{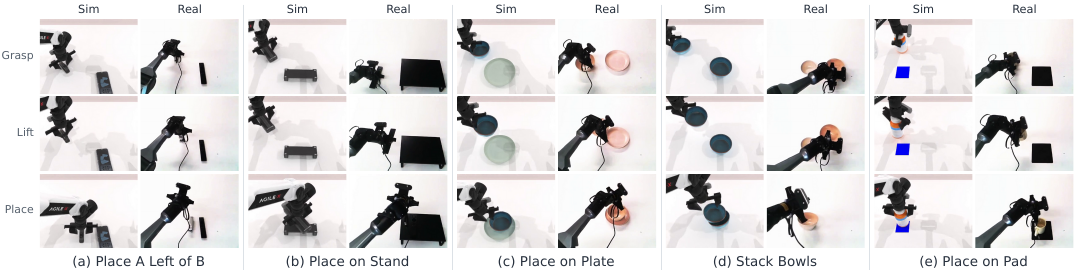}
\caption{\textbf{Cross-embodiment real-robot replay of simulation-generated trajectories.} For five RoboTwin task families, the policy runs in simulation using RFW-R assets and its recorded trajectory is then executed on a physical robot. \emph{Sim} and \emph{Real} show the grasp, lift, and place stages in simulation and on hardware, respectively. The real-robot phase is trajectory replay, not closed-loop policy inference.}
\label{fig:sim-real-vla}
\end{figure}

\paragraph{Cross-embodiment real-robot trajectory replay.}
To evaluate the physical executability of simulation-generated motions, we run the learned policies in RoboTwin scenes populated with RFW-R assets and record the resulting action trajectories. Since the simulated and physical robots have different embodiments, we retarget the corresponding end-effector (EEF) trajectories to the real robot for open-loop execution, rather than directly transferring joint-space commands. Inspired by simulation-to-hardware trajectory replay in \citet{wang2024penspin} and \citet{kappel2026simreplay}, we demonstrate real-robot execution across five task families. Figure~\ref{fig:sim-real-vla} presents paired simulation and real-robot frames illustrating the grasp, lift, and place stages. Policy inference is performed entirely in simulation, while the physical robot executes the retargeted trajectories without online policy feedback. These qualitative demonstrations are separate from the closed-loop simulation success rates reported in Table~\ref{tab:policy-transfer}. Additional implementation details are provided in Appendix~\ref{toc:G_policy3}.

\subsection{Scene construction and fulfillment}
\label{sec:exp-scene}
We evaluate the ability of RFW-R assets to satisfy task-driven scene specifications against EmbodiedGen v2 and ManiTwin. All three asset providers are tested on the same 16 fixed room foundations: one living room, two kitchens, seven offices, two bedrooms, three bathrooms, and one dining room. They share the requested objects, target sizes, support relations, and placement rules. In total, 265 requested instances form 153 groups, and unresolved requests remain in the fulfillment denominator. Shared camera target, azimuth, and distance candidates are used to select informative viewpoints. Five rendered views per provider--foundation pair yield 240 images across 48 conditions.

\begin{table}[htbp]\fontsize{8.5}{10}\selectfont
\caption{Task-driven scene fulfillment on 16 fixed room foundations. Inst. fill and Group fill report fulfilled requests out of 265 instances and 153 groups, respectively; Distinct is the number of unique assets used; Placement is the fraction of requested instances successfully placed.}
\label{tab:scene-fulfillment}
\centering

\begin{adjustbox}{max width=\linewidth}
\begin{tabular}{lrrrr}
\toprule
Library & Inst. fill \metricup & Group fill \metricup & Distinct \metricup & Placement \metricup \\
\midrule
\rowcolor{OursRow} \ours{} & \textbf{265/265} & \textbf{153/153} & \textbf{80} & \textbf{100\%} \\
EmbodiedGen v2 & 258/265 & 149/153 & 66 & 97.4\% \\
ManiTwin & 240/265 & 135/153 & 53 & 90.6\% \\
\bottomrule
\end{tabular}\end{adjustbox}
\end{table}
RFW-R fulfills all 265 requested instances and all 153 groups, compared with 258 instances and 149 groups for EmbodiedGen v2, and 240 instances and 135 groups for ManiTwin (Table~\ref{tab:scene-fulfillment}). It also places every requested instance under the shared placement rules and uses 80 distinct assets, indicating stronger coverage of the fixed task requirements than the compared libraries.

For visual assessment, a Qwen3.7-plus vision-language judge scores each provider--foundation condition on realism, visual functionality, layout, completeness, prompt alignment, and style. Each condition is judged three times at temperature zero; scores are first averaged within the condition and then across the 16 room foundations. We estimate paired mean-score differences using 10,000 bootstrap resamples of the foundations.

\begin{table}[htbp]\fontsize{8.5}{10}\selectfont
\caption{Visual assessment of scene construction across 48 provider--foundation conditions (0--10 scale). Visual func. denotes visual functionality, Complete denotes completeness, and Prompt denotes prompt alignment. Each condition is scored three times; scores are averaged within conditions and then across 16 foundations.}
\label{tab:scene-vlm}
\centering
\begin{adjustbox}{max width=\linewidth}
\begin{tabular}{lrrrrrrr}
\toprule
Library & Realism & Visual func. & Layout & Complete & Prompt & Style & Mean \\
\midrule
\rowcolor{OursRow} \ours{} & 3.52 & 4.19 & \textbf{4.88} & \textbf{3.02} & 3.27 & \textbf{4.40} & \textbf{3.878} \\
EmbodiedGen v2 & \textbf{3.67} & \textbf{4.25} & 4.71 & 2.92 & \textbf{3.38} & 4.15 & 3.844 \\
ManiTwin & 2.38 & 3.27 & 3.38 & 2.29 & 2.58 & 3.42 & 2.885 \\
\bottomrule
\end{tabular}\end{adjustbox}
\end{table}
RFW-R obtains the highest mean visual score (3.878), leading in layout, completeness, and style (Table~\ref{tab:scene-vlm}). Its advantage over ManiTwin is 0.993 points (95\% bootstrap interval [0.611, 1.444]). The difference relative to EmbodiedGen v2 is small (0.035 points; 95\% interval [$-0.108$, 0.191]), indicating comparable overall visual ratings. Combined with the fulfillment results, these findings highlight RFW-R's strength in supplying the requested objects and constructing task-aligned scenes.


\section{Conclusion}

We present RFW-W, a large-scale physical asset universe comprising over one million rigid, articulated, and deformable asset records across 2,228 dataset-specific category entries. By integrating category-guided asset generation, physical representations, and interaction annotations with task-driven scene construction, RFW-W provides a unified foundation for physically grounded embodied simulation. Extensive evaluations on RFW-R demonstrate broad asset coverage, reliable physical representations, and executable dexterous interactions. Furthermore, RFW-R achieves complete scene fulfillment in our evaluation and significantly improves downstream robotic manipulation performance over existing asset libraries. We believe RFW-W represents an important step toward scalable physical asset generation and the construction of diverse, interactive embodied worlds.

\FloatBarrier
\clearpage
\bibliographystyle{plainnat}
\bibliography{references}
\clearpage
\appendix
\renewcommand{\thesection}{A.\arabic{section}}
\renewcommand{\thesubsection}{\thesection.\arabic{subsection}}
\captionsetup[figure]{font=small}
\raggedbottom
\setcounter{table}{0}\setcounter{figure}{0}
\renewcommand{\thetable}{S\arabic{table}}
\renewcommand{\thefigure}{S\arabic{figure}}
\renewcommand{\theHtable}{supp.\arabic{table}}
\renewcommand{\theHfigure}{supp.\arabic{figure}}
\newcommand{\appentry}[2]{\par\vspace{3pt}\noindent\hyperref[#1]{\makebox[2.5em][l]{\textbf{\ref*{#1}}}\textbf{#2}}\nobreak\dotfill\hyperref[#1]{\pageref*{#1}}\par}
\newcommand{\appsubentry}[2]{\par\vspace{0pt}\noindent\hspace*{2.5em}\hyperref[#1]{#2}\nobreak\dotfill\hyperref[#1]{\pageref*{#1}}\par}
\section*{Appendix Contents}
\noindent\textit{Detailed methods, evaluation protocols, and supplementary results.}\par\vspace{7pt}
{\fontsize{9}{10.5}\selectfont\setlength{\parskip}{0pt}
\appentry{app:release}{Datasets and Evaluation Sets}
\appsubentry{toc:A_universe}{Universe census and counting conventions}
\appsubentry{toc:A_release_inventory}{Inventory and evaluation scope}
\appsubentry{toc:A_release0}{Dataset coverage}
\appsubentry{toc:A_release1}{Category-balanced evaluation}
\appsubentry{toc:A_release2}{Release scope and within-category diversity}
\appentry{app:repair}{Asset Construction Details}
\appsubentry{toc:B_repair0}{Generation and category reassignment}
\appsubentry{toc:B_repair_prompts}{Model and prompt specifications}
\appsubentry{toc:B_repair1}{Physicalization pipeline}
\appsubentry{toc:B_repair2}{Geometry cleanup}
\appsubentry{toc:B_repair3}{Recorded processing cost}
\appentry{app:audit}{Geometry and Semantic Evaluation}
\appsubentry{toc:C_audit0}{Geometry metrics}
\appsubentry{toc:C_audit1}{Semantic annotation metrics}
\appsubentry{toc:C_audit2}{Paired geometry and collision fidelity}
\appentry{app:physics}{Physical Representation and Composition}
\appsubentry{toc:D_physics0}{Part-level physical properties}
\appsubentry{toc:D_physics1}{Physical consistency}
\appsubentry{toc:D_physics3}{Effect of solid-fill modeling}
\appsubentry{toc:D_physics2}{Appearance, parts, and collision geometry}
\appentry{app:crosssim}{Cross-Simulator Evaluation}
\appsubentry{toc:E_crosssim0}{Drop-response protocol}
\appsubentry{toc:E_crosssim1}{Agreement metrics}
\appentry{app:grasp}{Dexterous Grasping: Protocol and Diagnostics}
\appsubentry{toc:F_grasp0}{Candidate generation and execution}
\appsubentry{toc:F_grasp1}{Candidate budgets}
\appsubentry{toc:F_grasp2}{Robustness and closure displacement}
\appentry{app:policy}{Policy Data, Training and Evaluation}
\appsubentry{toc:G_policy0}{Data budget and asset selection}
\appsubentry{toc:G_policy1}{Training and evaluation settings}
\appsubentry{toc:G_policy2}{Aggregation and task-level results}
\appsubentry{toc:G_policy3}{Cross-embodiment real-robot trajectory replay}
}
\clearpage
\FloatBarrier
\section{Datasets and Evaluation Sets}\label{app:release}
\subsection{Universe census and counting conventions}\label{toc:A_universe}
Table~\ref{tab:universe-inventory} records the inventories used to describe RFW-W. RFW-R contributes its retained rigid release. RFW-A contributes the 402,410-asset release and 531 categories reported in its release comparison~\citep{rfwa2026}. RFW-S distinguishes 100,000 completed generated assets from 83,365 assets in the configuration-matched intersection of physical verification and runtime checking~\citep{rfws2026}. Its five families contain 168 soft-solid, 89 cloth-sheet, 63 shell, 94 garment, and 71 bag categories, summing to 485; all remain represented after both checks.

\begin{table}[htbp]\centering
\caption{Inventory of RFW-W asset sources. Asset totals sum source-dataset records without cross-source deduplication; category totals count dataset--category entries. Inclusion criteria are specified per source.}
\label{tab:universe-inventory}
\begin{tabularx}{\linewidth}{@{}lrrX@{}}
\toprule
Source / accounting & Assets & Categories & Inclusion criterion \\
\midrule
RFW-R & 524,611 & 1,212 & Retained rigid release \\
RFW-A & 402,410 & 531 & Articulated release~\citep{rfwa2026} \\
RFW-S (jointly verified) & 83,365 & 485 & Same-configuration physical and runtime pass \\
\midrule
\rowcolor{OursRow} RFW-W, dataset sum & \textbf{1,010,386} & \textbf{2,228} & RFW-R + RFW-A + RFW-S intersection \\
\bottomrule
\end{tabularx}
\end{table}

Verification follows each asset type: RFW-R uses rigid-body diagnostics and drop-response tests; RFW-A reports full-release kinematic checks and a 200-asset passive-rollout study; RFW-S combines physical acceptance under frozen Genesis PBD/MPM configurations with a separate runtime check. The evaluations below use rigid-asset cohorts with the denominators specified in Table~\ref{tab:scope}.

\subsection{Inventory and evaluation scope}\label{toc:A_release_inventory}
The RFW-R rigid-object dataset contains 524,611 assets across 1,212 leaf categories. Table~\ref{tab:release-inventory} compares it with EmbodiedGen v2~\citep{embodiedgenv2_2026}, PhysX-3D~\citep{physx3d2025}, and ManiTwin~\citep{manitwin2026}. We report entropy-equivalent coverage $N_{\mathrm{eff}}=\exp(-\sum_c p_c\log p_c)$ within each library's taxonomy, where $p_c=n_c/\sum_j n_j$. The balanced evaluation samples two eligible assets per category (2,424 objects), independently of simulation and manipulation outcomes.

\begin{table}[htbp]\fontsize{8.5}{10}\selectfont
\caption{Rigid releases used in the shared evaluation. $N_{\mathrm{eff}}$ is entropy-equivalent class coverage within each library taxonomy.}
\label{tab:release-inventory}
\centering
\begin{adjustbox}{max width=\linewidth}
\begin{tabular}{@{}lrrrr@{}}
\toprule
Dataset & Assets & Classes & $N_{\mathrm{eff}}$ & Med./class \\
\midrule
\rowcolor{OursRow} \ours{} & \textbf{524,611} & \textbf{1,212} & \textbf{905.1} & \textbf{356} \\
EmbodiedGen v2 & 4,168 & 200 & 139.2 & 11 \\
PhysX-3D & 32,047 & 450 & 7.8 & 1 \\
ManiTwin & 4,574 & 485 & 285.6 & 6 \\
\bottomrule
\end{tabular}\end{adjustbox}
\end{table}

The effective coverage of RFW-R is 905.1 out of 1,212 categories, compared with 7.8 out of 450 for PhysX-3D. Thus, the RFW-R inventory is distributed across many categories, whereas the PhysX-3D inventory is concentrated in a small subset of its taxonomy. The median of 356 assets per RFW-R category complements this coverage measure by describing typical within-category availability.

\subsection{Dataset coverage}\label{toc:A_release0}
The catalog combines broad category coverage with multiple instances per class.

\begin{figure}[!htbp]
\centering
\includegraphics[width=\linewidth]{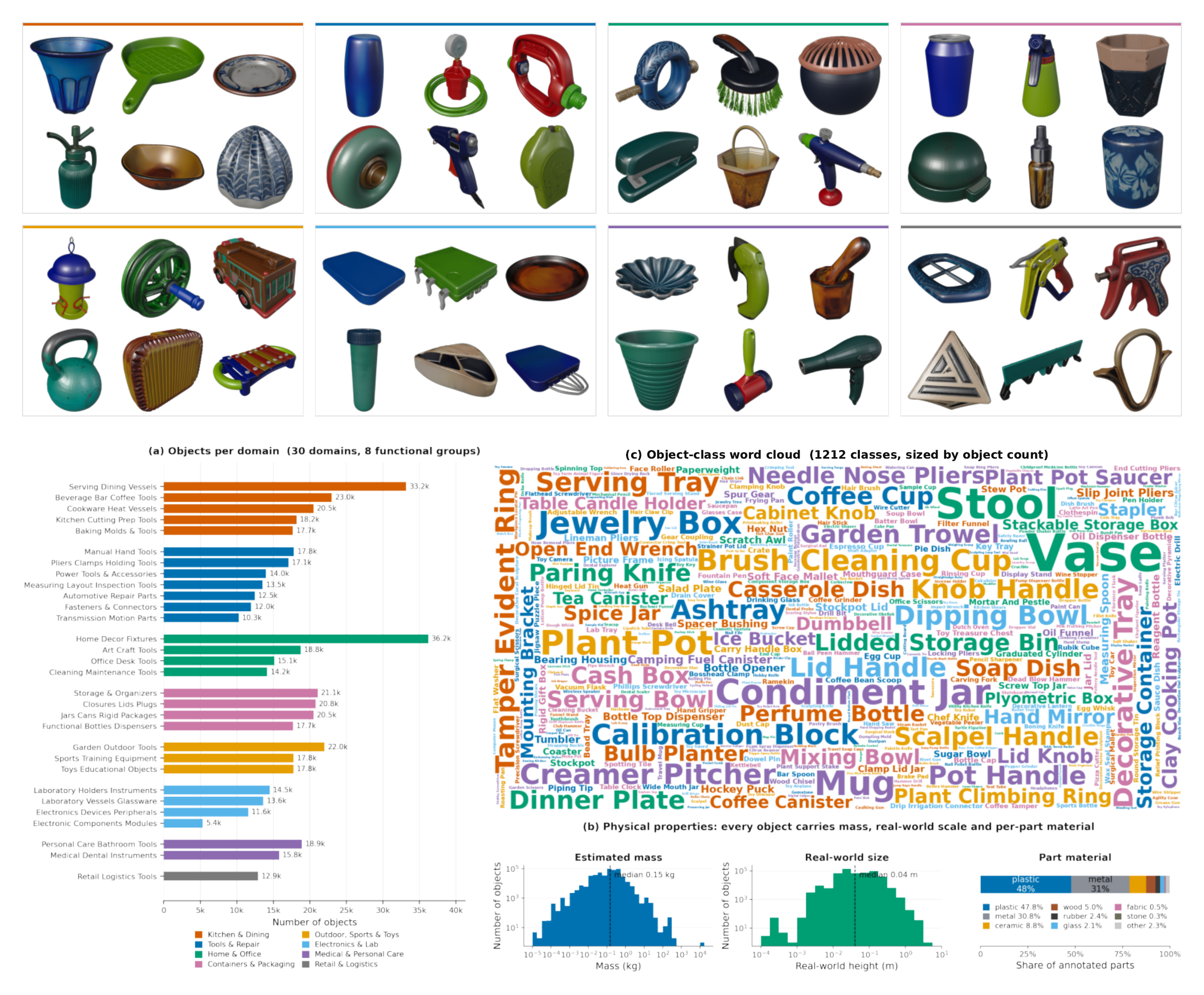}
\caption{\textbf{RFW-R dataset coverage.} Top: 48 representative assets from eight functional groups, individually scaled for visibility. Bottom: rounded object counts across 30 domains, a frequency-weighted word cloud of 1,212 categories, and distributions of estimated mass, metric height, and part materials. The final RFW-R release contains 524,611 retained assets across 1,212 categories.}
\label{fig:asset-overview}
\end{figure}

Figure~\ref{fig:asset-overview} summarizes the breadth of the rigid catalog with examples from eight functional groups and descriptive distributions across domains, classes, and estimated physical properties.

\FloatBarrier

\subsection{Category-balanced evaluation}\label{toc:A_release1}
We select two objects from each category with available geometry and physical descriptions, giving 2,424 objects. Selection does not depend on simulation or manipulation outcomes. The categories span 30 domains and comprise 121 head, 364 middle, and 727 tail categories, contributing 242, 728, and 1,454 selected objects.

\begin{table}[htbp]\fontsize{8.5}{10}\selectfont\centering
\caption{Evaluation settings.}
\label{tab:scope}
\begin{tabularx}{\linewidth}{lrlX}\toprule
Evaluation & Size & Unit & Setting \\\midrule
Rigid inventory & 524,611 & assets & 1,212 categories \\
Category-balanced sample & 2,424 & assets & Two per category \\
Four-engine agreement & 2,257 & assets & Common evaluable subset \\
Policy training & 2,000 & demonstrations & 100 per source and task \\
Policy evaluation & 4,000 & episodes & 100 per policy and condition \\
Scene construction & 16 & foundations & 265 instances, three providers \\\bottomrule
\end{tabularx}\end{table}
\begin{figure}[!htbp]
\centering
\includegraphics[width=\linewidth]{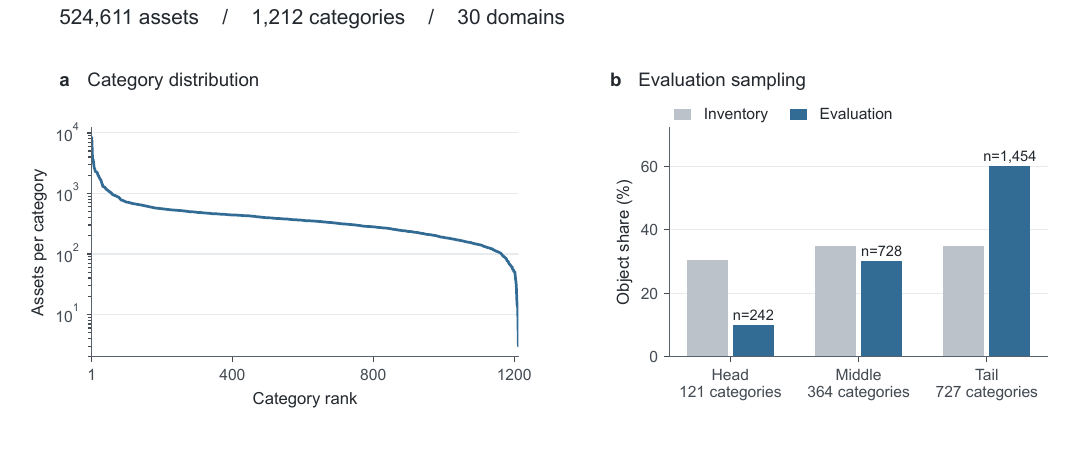}
\caption{\textbf{Rigid inventory and category-balanced sampling.} The category population curve uses a logarithmic vertical axis. Bars compare inventory and evaluation shares; $n$ labels give evaluation object counts. Sampling selects two objects per category.}
\label{fig:inventory-profile}
\end{figure}

\subsection{Release scope and within-category diversity}\label{toc:A_release2}
The retained inventory contains 524,611 assets; simulation export completes for 524,562 of them (99.99\%). The remaining 49 retained assets have no completed simulation export in this snapshot. All 2,424 category-balanced evaluation assets have complete simulation representations.

On the 2,424-object subset, an image-feature screen identifies 24 candidate near-duplicate clusters involving 55 objects (2.27\%) at cosine similarity 0.9782. The threshold is calibrated with same-category and cross-category proxy pairs to select candidates for visual inspection. Across the 1,212 within-category pairs, image-feature distance has median 0.1108 and 5th--95th percentiles 0.0525--0.1764.

These feature statistics describe variation within the balanced sample: only 2.27\% of sampled objects enter candidate near-duplicate clusters, while the within-category distances span a broad range. Candidate clusters are screening results for visual inspection, and the measured proportion applies to the 2,424 sampled objects.

\FloatBarrier
\section{Asset Construction Details}\label{app:repair}
\subsection{Generation and category reassignment}\label{toc:B_repair0}
Image generation, background removal, mesh reconstruction, five-view inspection, and category reassignment produce the retained inventory. The visual rubric checks single-object content, rigid-body suitability, completeness, floating fragments, recognizability, and category agreement. Objects failing only category agreement are reassigned through lexical and image-conditioned matching, followed by reinspection for breakage and multi-object assemblies. Suitable unmatched objects form new size-bounded classes.

\begin{table}[htbp]\fontsize{8.5}{10}\selectfont\centering
\caption{Candidate generation and retention after visual filtering and category reassignment.}
\label{tab:retention}
\begin{adjustbox}{max width=\linewidth}\begin{tabular}{lr}\toprule
Stage or disposition & Assets \\\midrule
Generated candidates & 599,439 \\
Retained with original label & 357,117 \\
Retained after reclassification & 147,573 \\
Retained in 12 induced categories & 19,921 \\
Total retained & 524,611 \\
Not retained in this inventory & 74,828 \\\bottomrule
\end{tabular}\end{adjustbox}\end{table}

Category reassignment retains 167,494 candidates through relabeling or new-category induction, accounting for 31.9\% of the retained inventory. This separates semantic mismatch from unusable geometry: visually suitable objects can contribute to coverage after their labels are corrected, while 74,828 candidates remain outside the retained inventory.

\subsection{Model and prompt specifications}\label{toc:B_repair_prompts}
\paragraph{Category-conditioned generation (L0).}
Qwen3.7-plus receives the domain and the English and Chinese category names and produces batches of English object descriptions in JSON. The prompt requests one sentence of 15--30 words per description, varying object sub-form, material and finish, style and era, color and pattern, condition and wear, and notable geometry. Twelve rotating diversity cues specify combinations of materials, styles, conditions, and colors. Generation uses temperature 0.9 and top-$p$ 0.95. Descriptions are deduplicated by token-set Jaccard similarity at 0.85 and screened for domain-specific material plausibility. FLUX.2-klein renders images from the selected descriptions with six inference steps and guidance scale 1.0; the supplementary Wan2.7 path samples descriptions across each category.

\paragraph{Visual quality and category prompts (L2).}
Qwen3.7-plus~\citep{qwen37plus} receives front, back, left, right, and bottom renders. Its ordered rubric asks whether the asset is a single object, a rigid body, complete, free of floating fragments, recognizable, and consistent with the requested domain and category. The response records these decisions and an object guess used for category reassignment. Ambiguous assignments are resolved from the rendered views and a candidate-category list, with an abstention choice. Follow-up prompts score visible breakage and the number of separate objects in a render.

\paragraph{Physical annotation prompt (L4).}
Qwen3.7-plus receives aligned textured RGB and color-coded part-mask views. The prompt uses the mask colors to locate parts and the RGB views to identify material category and subtype. For each part, it requests static and dynamic friction against rubber; for the complete rigid object, it requests metric dimensions and a mass estimate. The prompt asks for one answer per mask color and treats small or occluded parts in the context of the whole object. Material densities and part volumes then determine the composed mass, center of mass, and inertia in Appendix~\ref{app:physics}.

\subsection{Physicalization pipeline}\label{toc:B_repair1}
Figure~\ref{fig:pipeline} connects three physicalization stages. L3 takes the accepted mesh and produces semantic part labels and convex collision components. These serve distinct roles: parts index material and interaction annotations, while collision components specify contact geometry. L4 takes aligned RGB views and part masks, estimates physical attributes, and composes mass, center of mass, and inertia in a common object frame (Appendix~\ref{app:physics}). Assets are exported after completing L4, preserving geometry and physical quantities across simulation formats. L5 denotes downstream simulation evaluations of these exported assets. Evaluation outcomes provide diagnostic feedback for physical refinement. The cross-simulator evaluation uses the explicitly defined bounded-motion and penetration protocol in Appendix~\ref{app:crosssim}.

\subsection{Geometry cleanup}\label{toc:B_repair2}
Conservative cleanup targets coincident vertices, fragmented components, and inconsistent part annotations while preserving openings and graspable structures. Category reassignment and geometric cleanup serve different purposes: the former recovers category coverage, while the latter improves the representation used for annotation and export. A paired audit of the construction stages finds that visual surfaces remain essentially unchanged after coordinate alignment; Appendix~\ref{toc:C_audit2} reports the measurements. Physical normalization aligns the physical properties and geometry in a common object frame while preserving the visual surface.

\begin{figure}[!htbp]
\centering
\includegraphics[width=\linewidth]{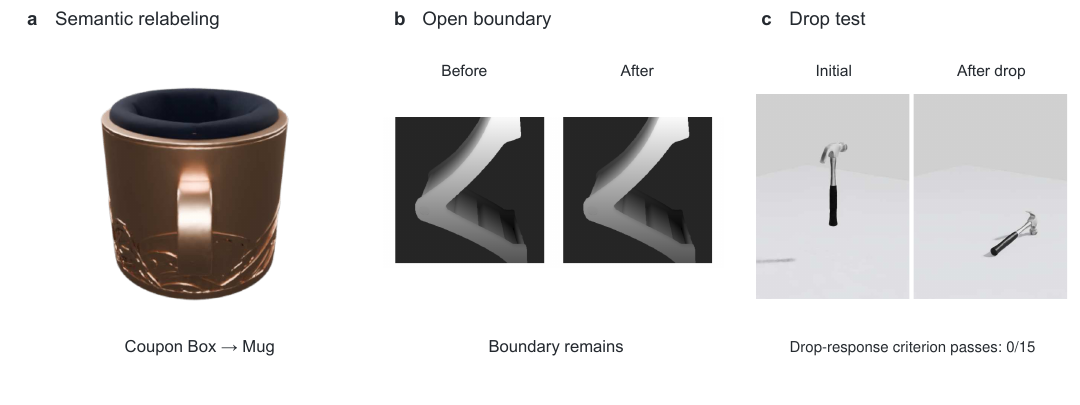}
\caption{\textbf{Observed diagnostic cases.} A source category is reassigned from Coupon Box to Mug; a boundary remains open after processing; and a claw hammer fails the specified drop-response criterion in all 15 archived drop trials. The panels show three independent diagnostic examples.}
\label{fig:observed-diagnostics}
\end{figure}

Figure~\ref{fig:observed-diagnostics} separates three diagnostic outcomes. Reassigning Coupon Box to Mug corrects semantic indexing. A remaining open boundary demonstrates that conservative cleanup preserves visual topology rather than guaranteeing a closed volume; volume-based measurements therefore require the eligibility check in Appendix~\ref{toc:C_audit2}. The hammer passes none of its 15 archived drop-response trials, identifying an asset that consistently violates at least one response criterion. This aggregate result identifies a failure under the downstream evaluation protocol; the violated criterion and its physical cause require trial-level diagnostics.

\subsection{Recorded processing cost}\label{toc:B_repair3}
Table~\ref{tab:processing-cost} reports CPU time for physical normalization, repair, and collision reconstruction. Each stage is measured independently on the asset set listed in the table.

\begin{table}[!htbp]\centering
\caption{Recorded asset-processing cost. CPU hours denote accumulated compute time. Asset counts correspond to the available timing records for each stage.}
\label{tab:processing-cost}
\begin{adjustbox}{max width=\linewidth}\begin{tabular}{lrrr}\toprule
Stage & Recorded assets & CPU hours & Mean seconds/asset \\\midrule
Physical normalization & 524,611 & 10,723.2 & 73.6 \\
Repair & 2,424 & 35.4 & 52.6 \\
Collision reconstruction & 1,291 & 358.8 & 1,000.5 \\
\bottomrule\end{tabular}\end{adjustbox}
\end{table}

The timing cohorts differ because the table summarizes available stage-specific records: normalization covers the retained inventory, repair covers 2,424 assets, and collision reconstruction covers 1,291. Collision reconstruction has the largest mean CPU cost per recorded asset (1,000.5\,s), identifying decomposition as a substantial cost within that measured cohort. These accumulated CPU hours describe compute consumption, not end-to-end wall-clock latency.

\FloatBarrier
\section{Geometry and Semantic Evaluation}\label{app:audit}
\subsection{Geometry metrics}\label{toc:C_audit0}
We characterize connected components, boundary-edge ratio, visual volume relative to convex-hull volume, and visual-to-collision surface distances. Deep intrusion measures collision samples lying more than 1\,cm inside the visual volume and is evaluated on watertight meshes. The original diagnostic settings use a component threshold of 5,000, boundary ratio 0.30, solidity 0.20, visual-to-collision P95 distance 2\,cm, and intrusion fraction 0.02.

The baseline conversion for PhysX-3D and ManiTwin uses visual geometry for collision, so its visual-to-collision distance is a self-comparison. Geometry scores therefore depend on the collision representation as well as the source asset. Table~\ref{tab:geometry-legacy} reports the diagnostic results for these source-specific collision representations.
\begin{table}[htbp]\fontsize{8.5}{10}\selectfont
\caption{Original geometry diagnostic pass rates (\%). Collision representations differ across sources as described in the text.}
\label{tab:geometry-legacy}
\centering

\begin{adjustbox}{max width=\linewidth}\begin{tabular}{lrrrr}
\toprule
Gate (pass rate \metricup) & \ours{} & EmbodiedGen v2 & PhysX-3D & ManiTwin \\
\midrule
Geometry-only strict & 79.3 & 71.1 & 53.7 & \textbf{89.5} \\
\midrule
Connectivity & 98.6 & 98.2 & 100 & 100 \\
Open-boundary edges & 98.9 & 96.1 & 99.9 & 100 \\
Visual$\rightarrow$collision dist. & 98.8 & 96.4 & 100 & 100 \\
Deep penetration & 99.5 & 100 & 100 & 100 \\
Hull solidity & 81.6 & 75.0 & 53.8 & \textbf{89.5} \\
\bottomrule
\end{tabular}\end{adjustbox}
\end{table}

Hull solidity is the lowest individual RFW-R pass rate (81.6\%), close to the joint geometry-only rate of 79.3\%; the other individual gates pass at least 98.6\%. Solidity is therefore the principal marginal bottleneck in this diagnostic. Low visual-volume-to-hull ratios can arise from concavities and openings as well as defective geometry, so this gate characterizes shape compactness under the chosen threshold. The 100\% visual-to-collision distance rates for PhysX-3D and ManiTwin follow from using the same geometry on both sides of that comparison.

\subsection{Semantic annotation metrics}\label{toc:C_audit1}
The four libraries share a Qwen3.7-plus vision-language scoring and aggregation protocol for category, language, functional, and physics-related agreement. CLIP~\citep{radford2021learning} image--text similarity provides an additional alignment measure. The physics-related judge score measures semantic plausibility and agreement with the estimated physical annotations. Qwen3.7-plus also assists annotation generation, so this score describes model-based annotation consistency; physical-parameter accuracy requires independent measurements. N/A indicates that the released comparison sources do not provide a comparable physics-judge field under this protocol.
\begin{table}[htbp]\fontsize{8.5}{10}\selectfont
\caption{Part-level annotation quality under the shared judge.
All four libraries use the same scoring and aggregation protocol.
$^{*}$EmbodiedGen v2 physics is computed on placeholder values; N/A denotes an unavailable comparable field.}
\label{tab:annotation-full}
\centering

\begin{adjustbox}{max width=\linewidth}\begin{tabular}{lrrrr}
\toprule
Dimension & \ours{} & EmbodiedGen v2 & PhysX-3D & ManiTwin \\
\midrule
Category \metricup & \textbf{0.981} & 0.93 & 0.96 & 0.65 \\
Language \metricup & 0.864 & 0.67 & \textbf{0.87} & 0.50 \\
Functional \metricup & \textbf{0.860} & 0.80 & 0.79 & 0.64 \\
Physics (judge) \metricup & \textbf{0.895} & 0.85$^{*}$ & N/A & N/A \\
\midrule
CLIP alignment \metricup & 0.244 & 0.232 & 0.238 & \textbf{0.261} \\
\bottomrule
\end{tabular}\end{adjustbox}
\end{table}

RFW-R obtains the highest category (0.981) and functional (0.860) scores, while ManiTwin has the highest CLIP alignment (0.261 versus 0.244). The metrics answer different questions: the shared judge evaluates specified annotation dimensions, whereas CLIP evaluates image--text embedding alignment. Their different rankings indicate that broad visual--textual alignment and task-relevant annotation agreement should be interpreted separately.

\subsection{Paired geometry and collision fidelity}\label{toc:C_audit2}
A paired audit tracks 2,424 assets through geometry cleanup and physical normalization. After coordinate alignment, the median per-object bidirectional surface-distance P95 is $3.87\!\times\!10^{-7}$\,m before and after physical normalization. Watertightness is approximately 10.1\% before and after normalization, consistent with preservation of the visual surface. Visual-mesh closure and simulator execution are separate properties: contact is handled by the collision representation, while a closed visual-volume proxy is required for volume IoU.

A separate stratified sample of 200 assets compares the released collision geometry, standard CoACD, and a single convex hull. We compute volume IoU on the 25 objects with valid closed visual-volume proxies. The other 175 sampled objects lack a valid closed visual-volume proxy for this calculation, leaving 12.5\% of the sample eligible. On the same 25 eligible objects, the released representation achieves median IoU 0.924, compared with 0.913 for CoACD and 0.682 for a single hull (Table~\ref{tab:collision-fidelity}). The larger gap to a single hull is consistent with multi-hull geometry retaining concave structure. This comparison characterizes the closed-proxy subset; surface-distance diagnostics above provide complementary evidence for geometry without a valid volume proxy.

\begin{table}[!htbp]\centering
\caption{Collision fidelity on the same 25 measurable objects from a 200-asset sample. IoU compares each collision volume with the closed visual-volume proxy.}
\label{tab:collision-fidelity}
\begin{tabular}{lrr}\toprule
Collision representation & Measurable objects & Median volume IoU \metricup \\\midrule
\rowcolor{OursRow} Released representation & 25 & \textbf{0.924} \\
Standard CoACD & 25 & \underline{0.913} \\
Single convex hull & 25 & 0.682 \\
\bottomrule\end{tabular}
\end{table}

\paragraph{Annotation continuity.}
Affordance annotations and the inputs to semantic evaluation are unchanged before and after physical normalization for all 2,424 audited objects. This checks continuity during physical normalization.

\FloatBarrier
\section{Physical Representation and Composition}\label{app:physics}
\subsection{Part-level physical properties}\label{toc:D_physics0}
Each part associates metric geometry with estimated material, density, friction, and semantic attributes. The resulting quantities are composed physical annotations derived from geometry and material priors. Given part masses $m_p=\rho_pV_p$, the object mass and center of mass are
\begin{equation}
M=\sum_p m_p,\qquad c=\frac{1}{M}\sum_p m_pc_p.
\end{equation}
Part inertia tensors are combined in the object frame using the parallel-axis theorem. A whole-object mass estimate $\widehat M$ anchors all part masses by the common factor $\widehat M/M$ when $M/\widehat M\notin[0.5,2]$. Uniform scaling preserves the composed center of mass and relative part masses.

\begin{figure}[!htbp]
\centering
\includegraphics[width=\linewidth]{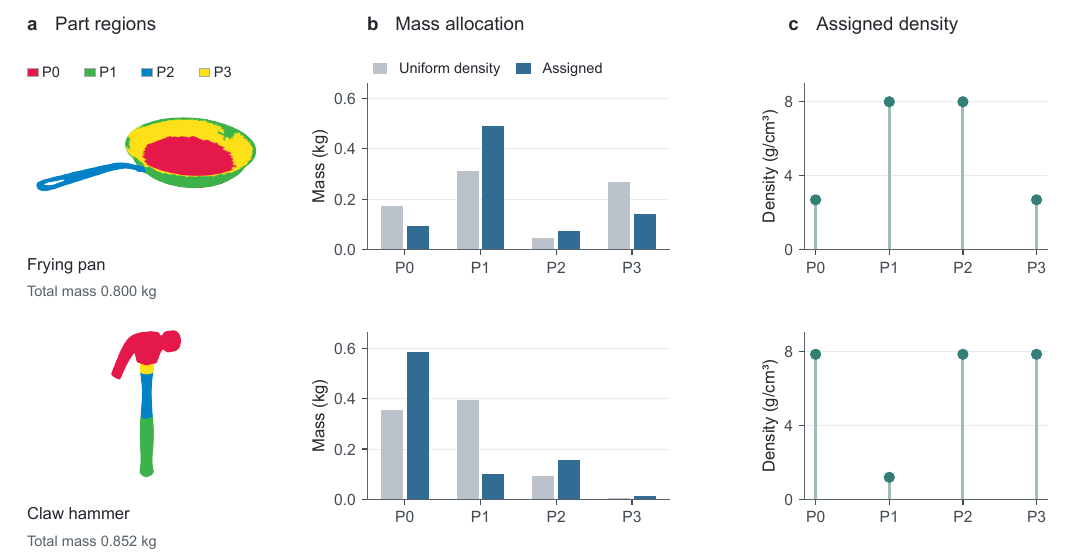}
\caption{\textbf{Part-wise physical assignment.} For each asset, the uniform-density baseline allocates the fixed total mass in proportion to stored part volume. Assigned masses use the exported annotations; both allocations sum to 0.800\,kg for the pan and 0.852\,kg for the hammer. The comparison shows how part annotations redistribute mass within the common object representation.}
\label{fig:physical-assignment}
\end{figure}

Figure~\ref{fig:physical-assignment} compares assigned part masses with a uniform-density allocation at the same total object mass. This isolates how the part annotations redistribute mass within each asset.

\subsection{Physical consistency}\label{toc:D_physics1}
We evaluate numerical consistency between stored physical annotations and simulator exports on 2,424 assets. The mass error below measures cross-format agreement, with stored/exported values as references rather than measured real-object mass. Object-level quantities are transformed with the geometry into a common frame; all audited center-of-mass and inertia tensors agree after alignment to the common object frame. Object mass recovered from URDF and MJCF has maximum relative error $1.5\!\times\!10^{-6}$. Part-mass sums agree with stored object mass within relative tolerance $10^{-3}$ for 98.5\% of assets. Preserving these physical quantities across simulator exports supports the reuse of RFW-R assets in scene construction and manipulation policy learning.

\begin{table}[!htbp]\centering
\caption{Physical-representation checks on 2,424 assets.}
\label{tab:physics}
\begin{adjustbox}{max width=\linewidth}\begin{tabular}{lr}\toprule
Check & Result \\\midrule
Simulation-description validity & 2,424/2,424 \\
Object-frame COM and inertia consistency & 2,424/2,424 \\
URDF/MJCF mass relative error (maximum) \metricdown & $1.5\!\times\!10^{-6}$ \\
Part-sum mass agreement within $10^{-3}$ \metricup & 98.5\% \\
\bottomrule\end{tabular}\end{adjustbox}
\end{table}

\FloatBarrier
\subsection{Effect of solid-fill modeling}\label{toc:D_physics3}
We compare the native representation with solid fill on 31 retained assets. One asset removed from the asset library because of a degenerate mass assignment is excluded from both conditions in every engine. The comparison uses MuJoCo, Genesis, and Isaac Sim, with three drop heights and five poses per asset, 2.5-second rollouts, and a 0.01-second time step. The two conditions yield 2,790 trials with no execution failures. Trial pass/fail labels follow the drop-response criteria in Appendix~\ref{app:crosssim}. Each asset contributes a criterion-pass fraction over 15 trials, and paired differences are aggregated across assets.

Table~\ref{tab:physics-ablation} isolates the solid-fill comparison. The native representation achieves higher criterion-pass fractions than solid fill in all three engines under the shared displacement, rotation, and penetration criteria. The estimated decreases range from 7.53 to 9.25 percentage points, and all three per-comparison intervals lie below zero. The common direction across engines supports sensitivity of the drop-response criterion to the representation change. A possible mechanism is that changing occupied volume and mass distribution alters contact response and rotational motion; identifying the contribution of geometry versus inertia requires the exact controlled-variable specification.

\begin{table}[!htbp]\centering
\caption{Effect of solid fill on 31 retained assets. Changes are solid fill minus native in criterion-pass fraction, expressed in percentage points. Approximate per-comparison 95\% intervals are computed as the paired mean $\pm1.96$ standard errors across assets.}
\label{tab:physics-ablation}
\begin{tabular}{lrr}\toprule
Engine & Mean change (pp) & Approx. 95\% interval \\\midrule
MuJoCo & $-9.25$ & $[-16.01, -2.48]$ \\
Genesis & $-9.25$ & $[-15.88, -2.62]$ \\
Isaac Sim & $-7.53$ & $[-13.52, -1.54]$ \\
\bottomrule\end{tabular}
\end{table}

\FloatBarrier
\subsection{Appearance, parts, and collision geometry}\label{toc:D_physics2}
\begin{figure}[!htbp]
\centering
\includegraphics[width=\linewidth]{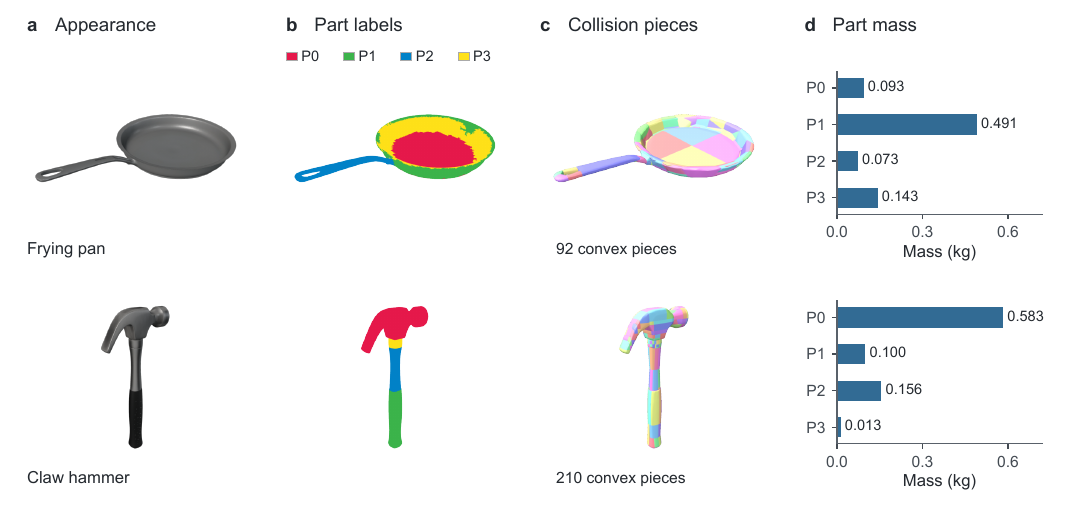}
\caption{\textbf{Aligned asset representations.} Appearance, original part labels, convex collision pieces, and assigned masses are shown for a frying pan and claw hammer. Part colors identify P0--P3; collision colors distinguish convex pieces. The masses are exported estimates. The illustrated part-based decompositions are distinct from the whole-object collision decompositions used in the quantitative experiments.}
\label{fig:asset-anatomy}
\end{figure}

A single rigid body can contain multiple semantic parts. Figure~\ref{fig:asset-anatomy} aligns the appearance, part segmentation, and collision representation of two assets. The frying pan and claw hammer each contain four annotated parts, with 92 and 210 convex hulls, respectively, in these illustrative part-based decompositions. The quantitative experiments use whole-object collision decomposition: semantic parts organize material and interaction attributes, while object-level convex components define simulation contacts. Thus, the part-based hull counts in this illustration describe the visualized examples rather than the collision construction evaluated quantitatively.

\FloatBarrier
\section{Cross-Simulator Evaluation}\label{app:crosssim}
\subsection{Drop-response protocol}\label{toc:E_crosssim0}
We evaluate the same 2,257 assets in all four engines. Each object is evaluated from heights 0.05, 0.15, and 0.30\,m with five initial orientations. The rollout lasts 2.5 seconds with a 0.01-second step. A valid trial passes the drop-response criterion when horizontal displacement is at most $\max(0.05\,\mathrm{m},0.1d)$, rotation change at most $8^{\circ}$, and support penetration no deeper than 0.01\,m, where $d$ is the object size. All engines use the estimated physical properties and multi-hull collision geometry. These fixed thresholds define a shared bounded-motion and penetration test: an object that tips or moves beyond a threshold receives a negative label even if its motion is physically plausible. Applying identical criteria makes engine responses comparable under this protocol.

\subsection{Agreement metrics}\label{toc:E_crosssim1}
For asset $i$ and engine $e$, let $s_{ie}$ be the fraction of valid trials that pass all three drop-response criteria and $b_{ie}=\mathbb{1}[s_{ie}\geq0.5]$. Pairwise agreement compares $b_{ie}$; Cohen's $\kappa$ corrects for marginal label frequencies. Fleiss' $\kappa$ summarizes four-engine agreement, and Pearson correlation uses the continuous values $s_{ie}$.

All pairs use the same 2,257 assets. All-engine agreement is 80.77\% with Fleiss' $\kappa=0.481$. Positive asset labels ($b_{ie}=1$) account for 7.75--13.03\% across engines. Agreement counts both matching positive and matching negative labels. Let $q_+$ and $q_-$ denote the fractions receiving positive and negative labels, respectively, in all four engines. Then $q_++q_-=80.77\%$. Since $q_+$ cannot exceed the smallest engine-positive fraction (7.75\%), $q_-\geq73.02\%$, up to rounding. Thus, most unanimous labels are negative under the bounded-motion criterion. The exact shared-positive and shared-negative fractions require the joint asset labels; the reported marginals establish these bounds. Fleiss' $\kappa$ complements raw agreement by accounting for the imbalanced label frequencies, while continuous-score correlations quantify whether engines rank the response fractions similarly.

\begin{figure}[htbp]\centering
\includegraphics[width=.92\linewidth]{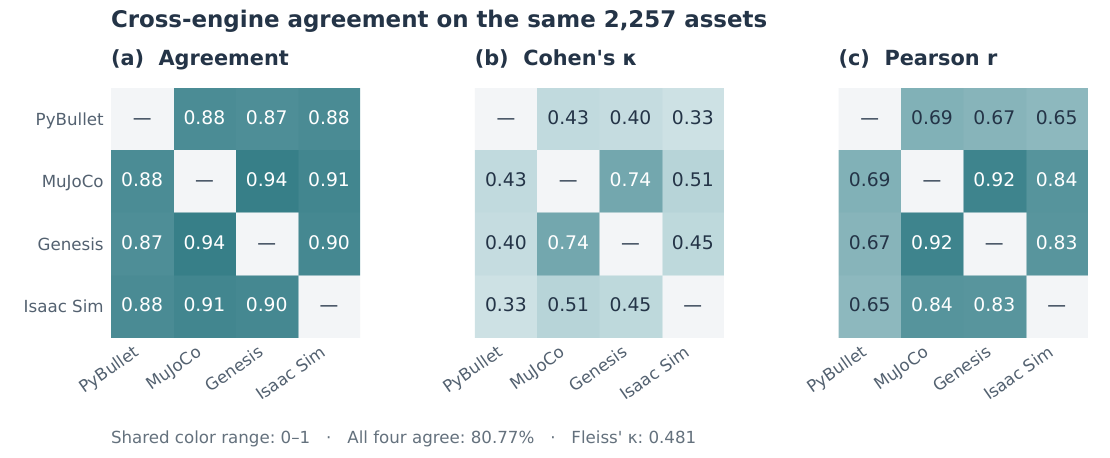}
\caption{\textbf{Drop-response consistency across four physics engines.} Pairwise agreement, Cohen's $\kappa$, and criterion-pass-fraction correlation use the same 2,257 assets. Values are calculated from the evaluated trials.}
\label{fig:crosssim-matrices}\end{figure}

\begin{figure}[htbp]
\centering
\includegraphics[width=\textwidth]{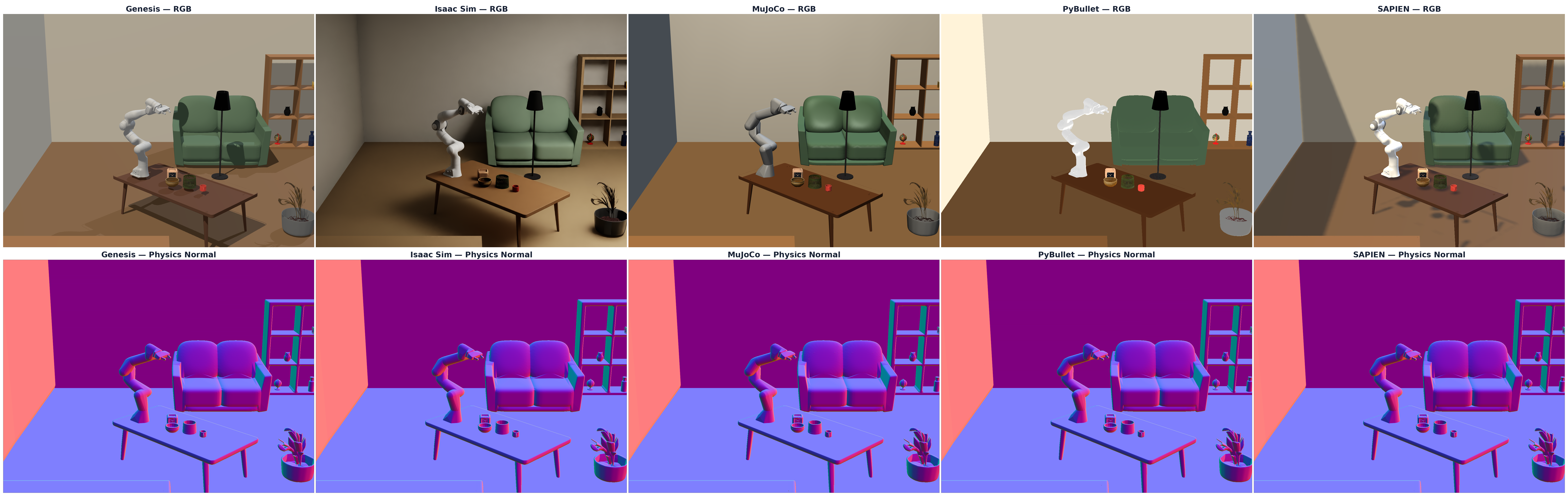}
\caption{\textbf{Cross-simulator scene rendering.} The same composed scene is rendered in five physics engines through the shared asset interface. Top: RGB views. Bottom: surface-normal views. These views illustrate scene structure and engine-dependent appearance through the shared interface. Task--scene correspondence is shown separately in Figure~\ref{fig:scene-multisim}; the five-engine rendering display is a qualitative interface demonstration, and behavioral agreement is evaluated separately on four engines by the drop-response protocol above.}
\label{fig:scene-multisim-normals}
\end{figure}

\FloatBarrier
\section{Dexterous Grasping: Protocol and Diagnostics}\label{app:grasp}
\subsection{Candidate generation and execution}\label{toc:F_grasp0}
We build on T(R,O) Grasp (TRO-Grasp)~\citep{fei2025trograspefficientgraph},
which generates hand-conditioned object--robot transformations. For each
target asset, we retain the generated pose representation and condition
the candidate pool on the selected semantic part through the partial
point-cloud interface. Candidates are converted to the common asset frame,
checked against the semantic part and real mesh, and then passed to the
same execution protocol for every trained hand. Up to 64 candidates per
object--hand pair are evaluated. The hand pose comprises root
translation/orientation and joint configuration.

Genesis executes closing, a 0.15\,m commanded lift, holding and
six-direction perturbation under open-loop position control. Lift-only
success requires object rise greater than 0.075\,m. Object-level
any-success counts an object as successful when at least one of its
evaluated candidates succeeds. The robust records combine
three mass/friction/pose settings. Cohort sizes are specified in
Table~\ref{tab:grasp-detailed}.

\begin{figure}[!htbp]
\centering
\includegraphics[trim=18 18 18 18,clip,width=\linewidth]{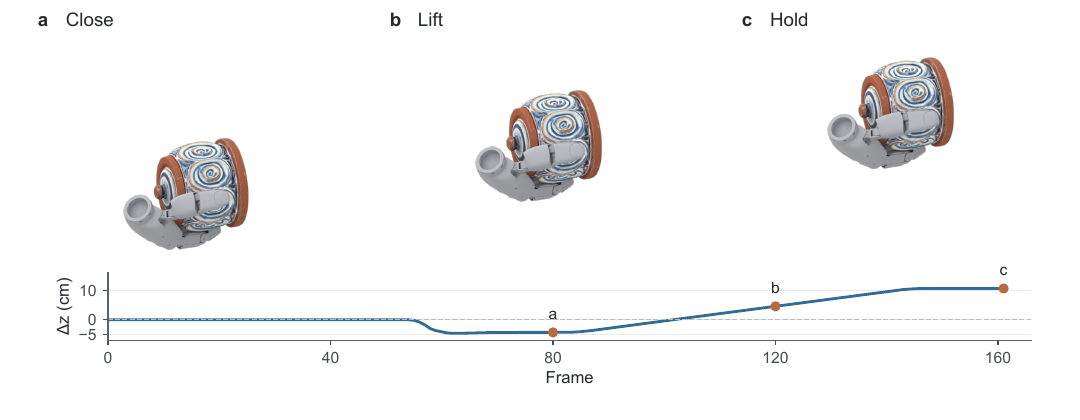}
\caption{\textbf{Example grasp execution.} XHand closes, lifts, and holds a condiment jar in three saved poses. Markers a--c locate frames 80, 120, and 161 on the vertical-displacement curve, measured relative to frame 0. The sequence illustrates the close, lift, and hold stages of object manipulation.}
\label{fig:grasp-sequence}
\end{figure}

A saved execution example in Figure~\ref{fig:grasp-sequence} shows the resulting close--lift--hold sequence.

\FloatBarrier
\subsection{Candidate budgets}\label{toc:F_grasp1}
The candidate budget controls how much of the generated grasp distribution is exposed to execution. Figure~\ref{fig:grasp-budget} shows the seven recorded budgets: coverage rises as the budget grows, with the largest gains for several hands occurring between 16 and 64 candidates. Table~\ref{tab:grasp-detailed} gives the full-budget rates and the separate robust cohorts. Increasing the maximum from 16 to 64 permits four times as many candidate executions per object--hand pair, exposing the coverage--validation-cost tradeoff. These are object-level any-success rates after candidate search, so the candidate budget is part of the evaluation protocol. At the full budget, Ability and Inspire exceed 92\% lift coverage, while Shadow reaches 42.3\%, showing substantial hand-dependent variation under the shared asset and execution pipeline.

\begin{table}[H]\fontsize{8.5}{10}\selectfont
\caption{Dexterous grasp outcomes in Genesis at the full candidate budget. Lift and robust rates use separate object cohorts ($N_L$, $N_R$). Brackets give Wilson 95\% intervals; macro averages include all six hands. The difference between lift and robust macro rates is not a paired perturbation effect because the cohorts differ.}
\label{tab:grasp-detailed}
\centering

\begin{adjustbox}{max width=\linewidth}\begin{tabular}{@{}lrlrr@{}}
\toprule
Hand & $N_L$ & Lift [95\% CI] \metricup & $N_R$ & Robust \metricup \\
\midrule
Ability & 2,372 & 92.5\% [91.3, 93.4] & 2,367 & 90.4\% \\
Inspire & 2,334 & 92.3\% [91.1, 93.3] & 2,353 & 90.4\% \\
Allegro & 2,372 & 91.3\% [90.1, 92.4] & 2,372 & 81.6\% \\
Barrett & 2,353 & 84.4\% [82.8, 85.8] & 2,372 & 81.5\% \\
XHand & 2,315 & 74.1\% [72.3, 75.8] & 2,315 & 71.2\% \\
Shadow & 2,353 & 42.3\% [40.3, 44.3] & 2,372 & 27.3\% \\
\midrule
Macro & & 79.5\% & & 73.7\% \\
\bottomrule
\end{tabular}\end{adjustbox}

\par\vspace{8pt}
\centering
\includegraphics[width=\linewidth]{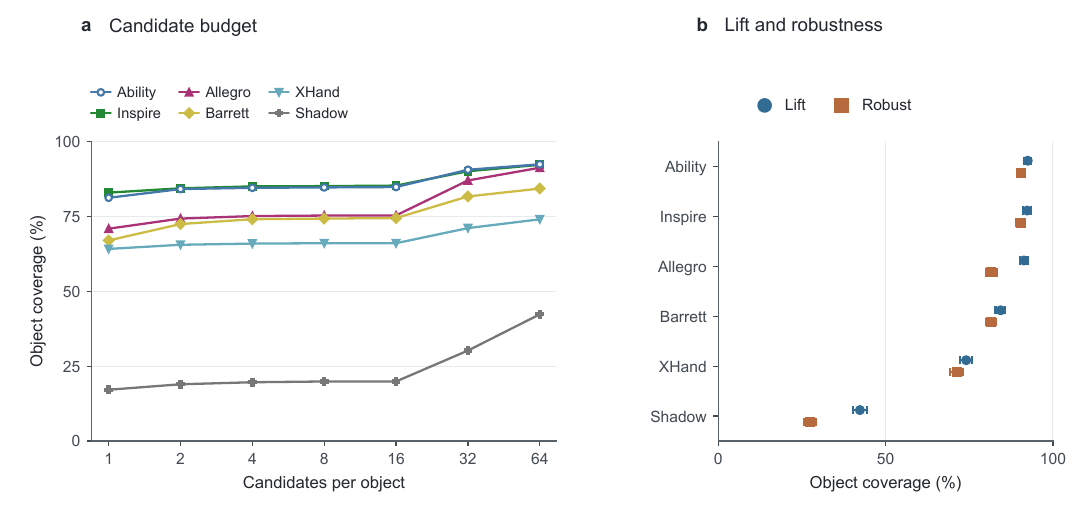}
\captionof{figure}{\textbf{Candidate budgets and execution outcomes.} Left: lift coverage at seven recorded budgets. Right: full-budget lift and robust coverage with Wilson 95\% intervals, using the separate cohorts in Table~\ref{tab:grasp-detailed}. Lines connect recorded observations.}
\label{fig:grasp-budget}
\end{table}
\FloatBarrier
\subsection{Robustness and closure displacement}\label{toc:F_grasp2}
Robustness is evaluated after the initial lift by combining holding with perturbations in mass, friction, and pose. Table~\ref{tab:grasp-closure} shows how many robust executions also satisfy the closure-translation bound, while Table~\ref{tab:grasp-stress} reports nominal, stronger, and combined perturbation outcomes on a separate stress-test cohort. Across the six hands, only five of the 10,432 robust successes fail the additional closure-translation bound. Closure displacement therefore removes few candidates that have already passed the robust test; this conditional count does not identify the causes of failed grasps. In the separate stress cohort, nominal-to-combined changes range from 1.0 to 5.9 percentage points, with the largest reduction for Allegro (86.3\% to 80.4\%). Ability and Inspire retain combined rates near 90\%, while Shadow remains below 41\% in all three settings. These outcomes distinguish baseline grasp coverage from sensitivity to the tested perturbations.

\begin{table}[htbp]\fontsize{8.5}{10}\selectfont\centering
\caption{Three-setting robust outcomes with an additional closure
translation bound that measures object displacement during gripper closure.}
\label{tab:grasp-closure}
\begin{adjustbox}{max width=\linewidth}\begin{tabular}{lrrrr}\toprule
Hand & N & Robust successes & Also bound & Also bound (\%) \metricup \\\midrule
Ability & 2,367 & 2,140 & 2,140 & 90.41 \\
Inspire & 2,353 & 2,126 & 2,125 & 90.31 \\
Allegro & 2,372 & 1,936 & 1,935 & 81.58 \\
Barrett & 2,372 & 1,934 & 1,933 & 81.49 \\
XHand & 2,315 & 1,648 & 1,648 & 71.19 \\
Shadow & 2,372 & 648 & 646 & 27.23 \\
\bottomrule\end{tabular}\end{adjustbox}\end{table}

\begin{table}[htbp]\fontsize{8.5}{10}\selectfont\centering
\caption{Separate stress-test cohort: nominal, stronger perturbation,
and combined stress rates. $N$ gives the number of evaluated objects for each hand.}
\label{tab:grasp-stress}
\begin{adjustbox}{max width=\linewidth}\begin{tabular}{lrrrr}\toprule
Hand & N & Nominal (\%) \metricup & Stronger (\%) \metricup & Combined (\%) \metricup \\\midrule
Ability & 2,372 & 91.6 & 91.1 & 90.6 \\
Inspire & 2,297 & 91.3 & 90.6 & 89.7 \\
Allegro & 2,372 & 86.3 & 80.7 & 80.4 \\
Barrett & 2,372 & 82.6 & 81.7 & 81.4 \\
XHand & 2,334 & 72.9 & 73.0 & 71.8 \\
Shadow & 2,353 & 40.8 & 40.0 & 38.7 \\
\bottomrule\end{tabular}\end{adjustbox}\end{table}

The same candidate record also stores semantic-part hits, penetration,
contact engagement, closure displacement, and physics outcomes, allowing
the three experiments above to use one asset-side annotation format.

\FloatBarrier
\section{Policy Data, Training and Evaluation}\label{app:policy}
\subsection{Data budget and asset selection}\label{toc:G_policy0}
For each of the four asset libraries, we train a separate policy on each of the five RoboTwin tasks using 100 successful demonstrations per task. This yields 20 policies trained on a total of 2,000 demonstrations, with the same demonstration budget across libraries.

Matching episode counts gives each source the same demonstration budget. The resulting comparison captures how the released asset coverage, trajectories, category match, and preparation choices combine to support policy learning. EmbodiedGen v2 uses one source asset per task, whereas the evaluation includes varied task objects, so the comparison measures the downstream utility of each released source under a fixed demonstration budget. Asset diversity and preparation remain components of this source-level comparison; attributing gains to an individual physicalization module requires a controlled ablation.

\subsection{Training and evaluation settings}\label{toc:G_policy1}
\textbf{Training.}
All four libraries use 100 demonstrations per task to fine-tune the $\pi_{0.5}$ base model with training seed 0 and global batch size 32. Evaluation uses the final checkpoint at step 19,999 for every source--task pair. Each source--task pair supplies its own training data and normalization statistics, and evaluation uses the final checkpoint without test-set selection.

\textbf{Closed-loop evaluation in simulation.}
Each of the 20 policies is evaluated on 100 clean and 100 randomized expert-solvable scenes, for 4,000 episodes. Clean and randomized suites are sampled independently. Randomization jointly varies visual context, table height, lighting, and instructions. Task success is the native RoboTwin success predicate; timeouts, planning failures, and infrastructure errors remain failures in each fixed denominator of 100. Each library has ten infrastructure-error episodes, all in the randomized condition.

The evaluation uses frozen suites. All libraries are evaluated using the same 100 seeds for each task--condition pair. The 100 evaluation episodes characterize each fixed trained checkpoint; training-seed variability is not estimated by this single-seed experiment.

\subsection{Aggregation and task-level results}\label{toc:G_policy2}
Macro averages weight the five tasks equally. The strongest baseline is ManiTwin in both modes: 55.4\% clean and 31.2\% randomized, compared with 71.2\% and 41.8\% for RFW-R. RFW-R leads on all five clean tasks and four randomized tasks; on randomized Move Pillbottle Pad, PhysX-3D reaches 9\% and RFW-R 6\%.

The clean-to-randomized macro difference for RFW-R is 29.4 percentage points. Plate and Bowls-2 retain 76\% and 73\% success, compared with 96\% and 98\% in clean scenes. A2B-L and Stand reach 21\% and 33\%, and Pillbottle-Pad drops from 52\% to 6\%, the largest task-level difference (46 points). Randomized Pillbottle-Pad is difficult for every source (0--9\%), identifying a shared task-level generalization gap. Because clean and randomized scenes are independently sampled and several factors change jointly, these differences summarize suite-level sensitivity; determining whether appearance, geometry, instructions, or planning dominates requires factor-wise evaluation or failure labels.

\subsection{Cross-embodiment real-robot trajectory replay}\label{toc:G_policy3}
We run the learned policies in RoboTwin simulation with RFW-R assets and retarget the resulting EEF trajectories to a physical Piper robot for cross-embodiment replay. The simulated and physical robots have different embodiments; EEF alignment provides the task-space correspondence for executing the recorded motions on hardware. Policy inference is performed entirely in simulation, and the real robot executes the retargeted sequence without online policy feedback from real-world observations.

\begin{figure}[htbp]
\centering
\includegraphics[width=\linewidth]{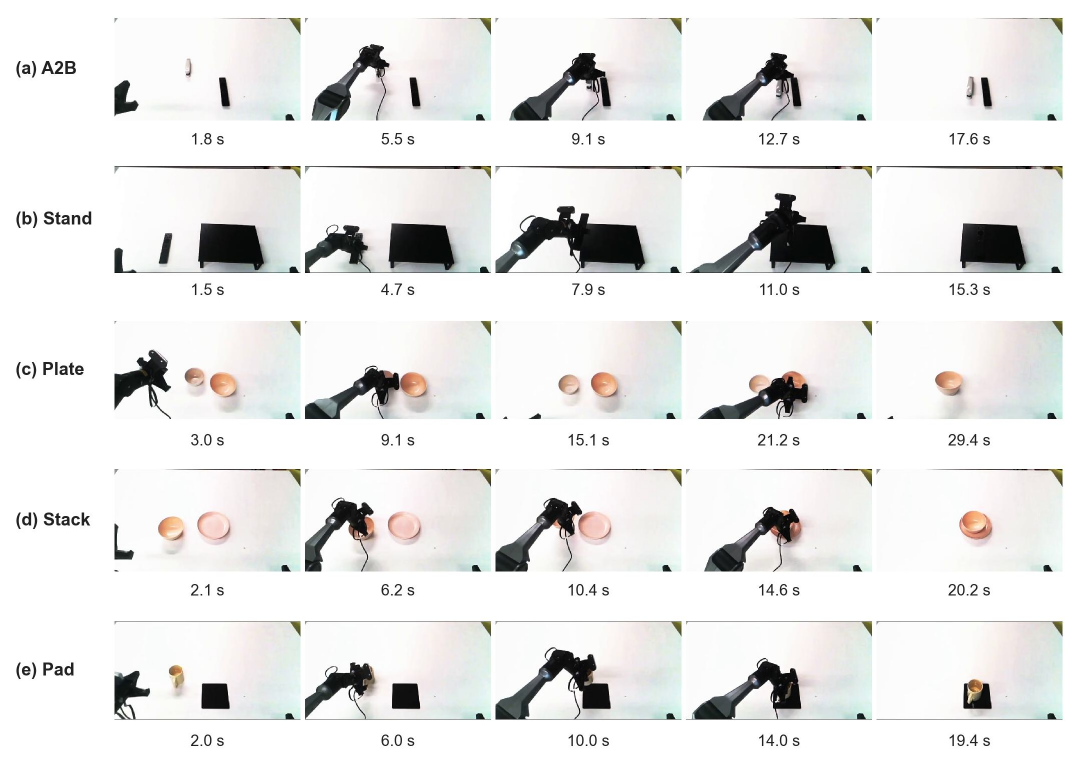}
\caption{\textbf{Temporal sequences of cross-embodiment real-robot trajectory replay.} The physical Piper robot executes EEF motions retargeted from policy trajectories generated in RoboTwin simulation. Rows show (a) place A left of B, (b) place on stand, (c) place on plate, (d) stack bowls, and (e) place on pad. Each row contains frames at 10\%, 30\%, 50\%, 70\%, and 97\% of its recorded replay video. The pad example uses a real mug and illustrates the placement task family.}
\label{fig:real-rollout-frames}
\end{figure}

Figure~\ref{fig:real-rollout-frames} shows grasp, lift, and placement stages across five task families. These sequences provide qualitative evidence of cross-embodiment motion execution. The pad example uses a real mug to illustrate the placement task family. Quantitative policy success rates in Table~\ref{tab:policy-transfer} are measured separately by closed-loop evaluation in simulation.

\FloatBarrier
\section{Future Work}

Future work will focus on three directions. First, we aim to extend part-level affordance and grasp annotations to articulated object generation and reconstruction, building on recent advances in articulated reconstruction and generation~\citep{liu2023paris,jiang2022ditto,weng2024digitaltwinart,liu2025artgs,liu2025videoartgs,shen2025gaussianart,wu2026reartgspp,yu2026part2gs,wang2025kinematify,jin2025uniart,chen2025artilatent,lu2025dreamart,shui2026articad,liu2025singapo,gao2025meshart,su2025artformer,zhang2026simart,wang2026artllm,yang2026sketch2arti}. 
Second, we plan to extend agentic annotation and scene generation to 3D editing, automatic rigging, and dynamic 3D/4D world generation, enabling scalable code-to-video data synthesis for embodied AI and autonomous driving~\citep{weng2026partflow,idea23d2025,zhang2025unirig,liu2025riganything,chen2026ovow,chen2026lumera,xu2025cruise,huang2026arthoi,li2026genncap,ren2023dreamgaussian4d,bahmani2024fourdfy,jiang2024consistent4d,xie2025sv4d,li2025ss4d,liu2025lightx,chen2026native4d,ye2025gsocc3d,wang2025unifying,zhang2024drone,zhang2026traffic,zhang2026dexora}.
Finally, we envision building a more comprehensive embodied benchmark and data engine by integrating diverse physical assets, interactive scenes, robot demonstrations, and multi-embodiment tasks~\citep{robotwin2_2025,liu2023libero,chen2026robodojo,cui2026liberosafety,zhang2026dexora,james2020rlbench,yu2020metaworld,mees2022calvin,gong2023arnold,mandlekar2022robomimic,walke2023bridgedatav2,khazatsky2024droid,zhang2025vlabench,fei2025liberoplus,geng2025roboverse}. This would enable systematic evaluation of manipulation capabilities, physical consistency, robustness, and sim-to-real generalization.
\end{document}